\documentclass[11pt]{article}

\newif\iffinalmode
\finalmodetrue
\usepackage[final]{acl}

\usepackage{times}
\usepackage{latexsym}
\usepackage{booktabs}
\usepackage{graphicx}
\usepackage{amsmath}
\usepackage{microtype}
\usepackage{xcolor}
\usepackage[T1]{fontenc}
\usepackage[utf8]{inputenc}
\usepackage{placeins}

\newcommand{\numFloresSents}{1,012}
\newcommand{\numSamayikTest}{2,417}
\newcommand{\numSamayikOod}{4,047}
\newcommand{\numItihasaTest}{11,721}
\newcommand{\numBootstrap}{1000}
\newcommand{\numBootstrapSeed}{0}
\newcommand{\numCiLevel}{95}
\newcommand{\numSaWords}{16,975}
\newcommand{\numEnWords}{21,901}
\newcommand{\numHiWords}{25,643}
\newcommand{\numParitySaEnLo}{1.774}
\newcommand{\numParitySaEnHi}{2.187}
\newcommand{\numParitySaEnGptTwo}{7.857}
\newcommand{\numParitySaHiLo}{1.060}
\newcommand{\numParitySaHiHi}{1.353}
\newcommand{\numParitySaHiLargeLo}{1.325}
\newcommand{\numParitySaHiLargeHi}{1.353}
\newcommand{\numFertSaLo}{3.11}
\newcommand{\numFertSaHi}{3.88}
\newcommand{\numFertSaGptTwo}{12.49}
\newcommand{\numFertSaGptTwoSlpOne}{3.97}
\newcommand{\numFertRatioSaEnLo}{2.31}
\newcommand{\numFertRatioSaEnHi}{8.38}
\newcommand{\numFertRatioSaHiLo}{1.60}
\newcommand{\numFertRatioSaHiHi}{1.74}
\newcommand{\numCompSaOrigLo}{1.63}
\newcommand{\numCompSaOrigHi}{7.21}
\newcommand{\numCompEnLo}{4.87}
\newcommand{\numCompEnHi}{4.92}
\newcommand{\numCompSaSlpLo}{2.05}
\newcommand{\numCompSaSlpHi}{2.38}
\newcommand{\numPreregFertLo}{5}

\newcommand{\numPreregParityThreshold}{1.5}
\newcommand{\numDeployedLo}{1.831}
\newcommand{\numDeployedHi}{2.899}
\newcommand{\numDeployedSamayikOTwoHundredK}{1.835}
\newcommand{\numDeployedSamayikOTwoHundredKCi}{[1.813, 1.858]}
\newcommand{\numTppBpeSixtyFourDeployed}{0.887}
\newcommand{\numTppBpeSixtyFourDeployedCi}{[0.875, 0.899]}
\newcommand{\numTppControlledBest}{1.035}

\newcommand{\numTppControlledBpeThirtyTwo}{1.084}
\newcommand{\numTppControlledSamayikLo}{1.030}
\newcommand{\numTppControlledSamayikHi}{1.142}
\newcommand{\numTppControlledOodLo}{1.055}
\newcommand{\numTppControlledOodHi}{1.163}
\newcommand{\numTppControlledFloresLo}{1.131}
\newcommand{\numTppControlledFloresHi}{1.219}

\newcommand{\numTppControlledItihasaAllLo}{0.555}
\newcommand{\numTppControlledItihasaAllHi}{0.658}
\newcommand{\numTppControlledHugeOodLo}{1.015}
\newcommand{\numTppControlledHugeOodHi}{1.145}
\newcommand{\numControlledPairs}{4}
\newcommand{\numControlledPairsAll}{12}
\newcommand{\numControlledPairsSmall}{8}
\newcommand{\numControlledPairsHuge}{4}
\newcommand{\numBmLines}{78,624}
\newcommand{\numBmLinePct}{67.7}
\newcommand{\numBytesSanskritTrain}{11.21}
\newcommand{\numBytesEnglishTrain}{16.55}
\newcommand{\numBytesEnglishBmTrain}{11.21}
\newcommand{\numBytesEnglishExcessPct}{48}
\newcommand{\numBmMoveCount}{24}
\newcommand{\numBmMaxMove}{0.025}
\newcommand{\numBmMedianMove}{0.008}
\newcommand{\numBmDownwardCount}{23}
\newcommand{\numBmVerdictChanges}{0}
\newcommand{\numVocabHuge}{128,000}
\newcommand{\numTppBpeHugeSamayik}{0.983}
\newcommand{\numTppBpeHugeSamayikCi}{[0.971, 0.997]}
\newcommand{\numTppBpeHugeSamayikBm}{0.978}
\newcommand{\numTppBpeHugeSamayikBmCi}{[0.965, 0.991]}
\newcommand{\numTppBpeHugeOod}{1.025}
\newcommand{\numTppBpeHugeOodCi}{[1.013, 1.037]}
\newcommand{\numTppBpeHugeOodBlockCi}{[0.999, 1.050]}
\newcommand{\numTppBpeHugeFlores}{1.116}
\newcommand{\numBpeMonotoneSequences}{7}
\newcommand{\numBpeSequencesTotal}{8}
\newcommand{\numBpeFloresStepGap}{0.001}
\newcommand{\numUnigramBmPieces}{50,659}

\newcommand{\numCharRatioSamayik}{1.028}
\newcommand{\numCharRatioItihasa}{0.596}
\newcommand{\numCharRatioFactor}{1.725}
\newcommand{\numDensityRatioSamayikLo}{1.002}
\newcommand{\numDensityRatioSamayikHi}{1.111}
\newcommand{\numDensityRatioItihasaLo}{1.001}
\newcommand{\numDensityRatioItihasaHi}{1.104}
\newcommand{\numDensityRatioSamayikHugeLo}{0.951}
\newcommand{\numDensityRatioSamayikHugeHi}{0.956}
\newcommand{\numDensityPairGap}{0.032}
\newcommand{\numTSevenSamayik}{1.027}
\newcommand{\numTSevenItihasa}{0.595}
\newcommand{\numTSevenVocab}{256}
\newcommand{\numBlockLength}{50}
\newcommand{\numBlockRowsPerCorpus}{13}
\newcommand{\numBlockWidenItihasaLo}{2.4}
\newcommand{\numBlockWidenItihasaHi}{2.6}
\newcommand{\numBlockWidenFloresLo}{2.1}
\newcommand{\numBlockWidenFloresHi}{2.6}
\newcommand{\numBlockWidenOodLo}{1.8}
\newcommand{\numBlockWidenOodHi}{2.4}
\newcommand{\numBlockWidenSamayikLo}{0.9}
\newcommand{\numBlockWidenSamayikHi}{1.0}
\newcommand{\numBlockNarrowerSamayik}{11}
\newcommand{\numBlockItihasaMaxUpper}{0.667}
\newcommand{\numEnTokensOTwoHundredK}{39,339}
\newcommand{\numEnTokensEOneBpe}{33,702}
\newcommand{\numEnTokenSavingPct}{14.3}
\newcommand{\numUnigramSixtyFourPieces}{62,896}

\newcommand{\numVocabMin}{50,257}
\newcommand{\numVocabMax}{262,145}
\newcommand{\numLargeVocabFloor}{200,019}
\newcommand{\numVocabSmall}{32,000}
\newcommand{\numVocabLarge}{64,000}
\newcommand{\numArmsDeployedGeneral}{4}
\newcommand{\numArmsDeployedIndic}{3}
\newcommand{\numTrainedArms}{6}
\newcommand{\numCheckedSentences}{19,197}
\newcommand{\numLeakedSentences}{0}
\newcommand{\numBytesPerDevanagariChar}{three}
\newcommand{\numSamayikTrain}{43,493}
\newcommand{\numItihasaTrain}{75,161}
\newcommand{\numTrainPairsTotal}{118,654}
\newcommand{\numRenyiAlpha}{2.5}
\newcommand{\numRenyiDeployedLo}{0.568}
\newcommand{\numRenyiDeployedHi}{0.581}
\newcommand{\numRenyiGptTwo}{0.270}
\newcommand{\numRenyiGptTwoSlpOne}{0.578}
\newcommand{\numRenyiDeployedSlpOneLo}{0.525}
\newcommand{\numRenyiDeployedSlpOneHi}{0.566}
\newcommand{\numRenyiBpeThirtyTwo}{0.615}
\newcommand{\numRenyiBpeSixtyFour}{0.589}
\newcommand{\numRenyiUnigramLo}{0.485}
\newcommand{\numRenyiUnigramHi}{0.493}
\newcommand{\numRenyiEnglishLo}{0.490}
\newcommand{\numRenyiEnglishHi}{0.508}
\newcommand{\numLengthSparseBelow}{30}
\newcommand{\numSamayikMeanEnWords}{12.5}
\newcommand{\numSamayikMeanSaWords}{9.6}
\newcommand{\numItihasaMeanEnWords}{30.7}
\newcommand{\numItihasaMeanSaWords}{11.2}
\newcommand{\numLengthGradientsFlipped}{16}
\newcommand{\numLengthGradientsTotal}{16}
\newcommand{\numLengthEnFirstBin}{1--8}
\newcommand{\numLengthEnFirstLo}{1.196}
\newcommand{\numLengthEnFirstHi}{1.304}
\newcommand{\numLengthEnLastBin}{25--40}
\newcommand{\numLengthEnLastLo}{0.934}
\newcommand{\numLengthEnLastHi}{1.026}
\newcommand{\numLengthPairsBelowOneEn}{3}
\newcommand{\numLengthSaFirstBin}{1--5}
\newcommand{\numLengthSaFirstLo}{0.894}
\newcommand{\numLengthSaFirstHi}{0.986}
\newcommand{\numLengthSaLastBin}{16--25}
\newcommand{\numLengthSaLastLo}{1.101}
\newcommand{\numLengthSaLastHi}{1.228}
\newcommand{\numLengthPairsBelowOneSa}{4}
\newcommand{\numVerseBelowProseComparisons}{28}
\newcommand{\numVerseBelowProseTotal}{28}
\newcommand{\numVerseProseGapMin}{0.117}
\newcommand{\numVerseProseGapMax}{0.541}

\title{Fewer Words, Not Fewer Tokens: Measuring the Sanskrit Tokenization Penalty per
Proposition}

\author{Devansh Sharma \\
  Independent researcher \\
  \texttt{sharmadevansh436@gmail.com}}

\begin{document}
\maketitle

\begin{abstract}
Sanskrit fuses case, number, person and tense into word endings and chains clauses into
compounds, so it is information-dense per word. Whether that density survives subword
tokenization is a separate question, to be asked per unit of meaning rather than per word.
On identical FLORES-200 devtest content, Sanskrit costs
\numParitySaEnLo--\numParitySaEnHi{} times the English tokens under deployed
tokenizers with vocabularies of \numLargeVocabFloor{} ids or more, but only
\numParitySaHiLargeLo--\numParitySaHiLargeHi{} times the Hindi tokens. Against a deployed
English tokenizer, Sanskrit-trained BPE arms then look \emph{cheaper} per proposition than
English on contemporary prose (\numTppBpeSixtyFourDeployed). Against a matched English
control, the same algorithm and vocabulary trained on the English side of the same corpus,
that flip disappears: at \numVocabSmall{} and \numVocabLarge{} pieces all
\numControlledPairsSmall{} matched pairs, each size-matched arm against both a pair-matched
and a byte-matched control, sit above 1.0 on prose with \numCiLevel\%{} intervals
excluding it. The gap closes as
the vocabulary grows: at \numVocabHuge{} pieces the BPE pair reads
\numTppBpeHugeSamayik{} in domain while staying above parity out of domain
(\numTppBpeHugeOod) and on FLORES (\numTppBpeHugeFlores). The ratio factorises into a
character-length ratio and a tokens-per-character ratio, the second near 1 throughout: what
survives matched tokenization is character-level length, which Sanskrit
prose lacks over English in SLP1 (\numCharRatioSamayik) and Sanskrit verse has
(\numCharRatioItihasa). The robust statement is about deployed practice: on
contemporary prose and on FLORES, with the Sanskrit side in SLP1 against the deployed
\texttt{o200k} English pivot, Sanskrit costs \numDeployedLo--\numDeployedHi{} English
tokens per proposition under the tokenizers people actually ship. Code, the results
snapshot and every table here are public.
\end{abstract}

\section{Introduction}
\label{sec:intro}

Two claims about Sanskrit are easy to run together, and this paper depends on keeping
them apart.

\textbf{Claim A} is that Sanskrit is information-dense per word. Its eight cases mark on
the noun what English marks with prepositions and word order; its verbs fuse person,
number, tense, mood and voice into one ending; and compounding chains what would be a
relative clause into a single orthographic word. We cite this claim rather than test it.
The idea that Sanskrit's grammatical tradition encodes structure a computational system
could use is old, and \citet{briggs-1985-knowledge} is its most-cited modern statement.

\textbf{Claim B} is that the density survives tokenization. This is the claim we test, and
it is not implied by Claim A. Fewer words is not fewer tokens. A subword learner takes
whitespace as its cheapest boundary signal, and sandhi, the obligatory fusion at word
boundaries, deletes exactly that signal. A tokenizer may spend its vocabulary rediscovering
structure the language had already marked, and any measurement that divides by a word
count will hide this, because Sanskrit's word count is small for the same reason its words
are long.

We measure the cost per unit of meaning on parallel text, where a Sanskrit sentence and
its English translation carry approximately the same content, and we add the control that
turns the measurement into a comparison: an English tokenizer trained with the same
algorithm, at the same vocabulary size, on the English side of the same corpus. Without it
the Sanskrit arm is compared against a tokenizer differing in vocabulary size and training
domain as well as language, and a crossing below parity is not attributable to language.
The scope is token cost and nothing downstream of it, so the modelling outcomes those
counts might predict are left to follow-up work.

Our contributions are three. \textbf{The measurement:} parity and tokens per proposition
for Sanskrit against English and Hindi on identical content, across
\numArmsDeployedGeneral{} deployed general-purpose tokenizers, \numArmsDeployedIndic{}
deployed Indic tokenizers and \numTrainedArms{} tokenizers trained here, with paired
bootstrap intervals on every tokens-per-proposition ratio, and with fertility reported
alongside and never as the headline. \textbf{The matched control:} a same-algorithm,
same-vocabulary, same-corpus English control family in two forms, one matched to the
Sanskrit training text sentence for sentence and one subsampled to its byte count, at
\numVocabSmall{}, \numVocabLarge{} and \numVocabHuge{} pieces. It removes the apparent
Sanskrit advantage on prose at the two smaller sizes under either form, and leaves it
standing at the largest on in-domain prose alone. The
Sanskrit token count is the same in both ratios by construction, so the whole difference
between them is the English denominator, and what is measured is that the control spends
\numEnTokenSavingPct\%{} fewer tokens on the same English text, because it was trained on
that domain. \textbf{The artefacts:} a public repository in which each experiment is one
command, every evaluation sentence is excluded from every training set by a hashed list on
both sides, and every table and figure here is regenerated from a tracked results snapshot
by a script.

\section{Related work}
\label{sec:related}

\paragraph{Comparing languages on parallel text.} Holding content constant and letting
the language vary is the design \citet{mielke-etal-2019-kind} use to ask which languages
are hard to model, and \citet{bugliarello-etal-2020-easier} make the same move
information-theoretically, finding translation out of English easier than into it: a
statement about direction Section \ref{sec:setup} has to take seriously. Whether languages
differ in how much they convey per unit is older than NLP.
\citet{pellegrino-etal-2011-cross} and \citet{coupe-etal-2019-different} measure
information rate in speech and find syllable rate and information density trading off, so
that rate per unit of time stays near constant; tokens per proposition is that
normalisation applied to a token budget rather than a channel.

\paragraph{Tokenizer inequity and its cost.} That a shared multilingual vocabulary spends
more tokens on some languages than others, at a cost in money and context window, is
established \citep{petrov-etal-2023-tokenizers,ahia-etal-2023-languages}, and
\citet{arnett-etal-2025-inequities} account for where the inequity comes from.
\citet{rust-etal-2021-good} show the tokenizer carrying much of the monolingual
performance gap in multilingual models, and \citet{limisiewicz-etal-2023-tokenization}
relate vocabulary allocation and overlap to downstream behaviour.

\paragraph{What tokenizer metrics predict.} Whether an intrinsic measure predicts anything
downstream is contested. \citet{bostrom-durrett-2020-byte} and
\citet{gowda-may-2020-finding} report vocabulary construction and size changing downstream
results and \citet{goldman-etal-2024-unpacking} find compression correlated with
performance, while \citet{schmidt-etal-2024-tokenization} argue compression is not the
whole account, \citet{ali-etal-2024-tokenizer} find fertility and parity unreliable
predictors of it, and \citet{uzan-etal-2024-greed} show inference-time segmentation
mattering alongside the vocabulary; R\'enyi efficiency
\citep{zouhar-etal-2023-tokenization} and its counterexamples
\citep{cognetta-etal-2024-counterexamples} belong to the same debate. We take its
conservative side and claim nothing about what a token count buys.

\paragraph{Translationese, and morphologically rich languages.} A translation differs
systematically from text originally written in the same language
\citep{koppel-ordan-2011-translationese,volansky-etal-2015-features}, and its direction
changes what an evaluation measures \citep{graham-etal-2020-statistical}; because tokens
per proposition divides by a translation, that bears on the sign of the bias in every
number here, which Section \ref{sec:setup} states corpus by corpus. On morphology,
\citet{klein-tsarfaty-2020-getting} ask whether word pieces suit complex morphology,
\citet{toraman-etal-2023-impact} measure tokenizer granularity against Turkish,
\citet{arnett-bergen-2025-language} attribute most of the language-modelling gap for
morphologically complex languages to training-set size once byte premiums are accounted
for, and \citet{arnett-etal-2025-morphscore} measure agreement between token and gold
morpheme boundaries in 70 languages.

\paragraph{What this paper adds.} Not a metric: Equation \ref{eq:tpp} is parity
\citep{petrov-etal-2023-tokenizers} with a different tokenizer on each side. What is added
is the matched denominator, and the separation of fertility from parity for a language
whose word boundaries sandhi erases.

\section{Background}
\label{sec:background}

Four features of Sanskrit matter for tokenization, and none of them is exotic once
stated in NLP terms.

\textbf{Sandhi} is euphonic combination at morpheme and word boundaries. \emph{tat} plus
\emph{api} is written \emph{tadapi}: the boundary is not merely unmarked, the segments
either side of it are altered. The rules are deterministic forwards and ambiguous in
reverse, which is why sandhi splitting is a research problem with its own literature
\citep{hellwig-nehrdich-2018-sanskrit,sandhan-etal-2022-translist,%
nehrdich-etal-2024-byt5} and datasets \citep{krishna-etal-2017-dataset}.

\textbf{Sam\=asa} is compounding: several stems fuse into one word and only the last
inflects, so a single orthographic word can carry what English writes as a noun phrase or
a relative clause. \textbf{Vibhakti} are the case endings, eight of them, encoding
grammatical role on the noun itself and replacing much of what English spends prepositions
and function words on.

\textbf{SLP1} is a lossless one-character-per-phoneme ASCII transliteration. Every metric
is computed on the original script and on SLP1 separately, because the choice is not
neutral: a vocabulary that already covers Devanagari gains nothing from romanisation, and
one that does not gains a great deal.

The consequence for measurement is the reason this paper exists. Fertility divides by a
denominator that sandhi and compounding make small, so a language writing in fewer, longer
words is penalised for the very property under test, and the penalty is arithmetic rather
than empirical. We report fertility for comparability with the tokenizer-inequity
literature \citep{petrov-etal-2023-tokenizers,ahia-etal-2023-languages,%
singh-etal-2024-indicgenbench,arnett-etal-2025-inequities} and never lead with it.

\section{Measurement}
\label{sec:measurement}

Let $T$ be a tokenizer, $s$ a sentence, $|T(s)|$ the number of tokens it produces, $w(s)$
the number of whitespace-delimited words in $s$, and $b(s)$ the number of UTF-8 bytes in
$s$. All four quantities below are ratios of sums over a corpus, not means of per-sentence
ratios, so a single long sentence contributes in proportion to its length, and the corpus
number is a length-weighted average. Section \ref{sec:length} is where that becomes
material.

\paragraph{Fertility and compression.} Fertility is tokens per word,
$\sum_{s} |T(s)| / \sum_{s} w(s)$, reported and never headlined. Compression is bytes per
token, $\sum_{s} b(s) / \sum_{s} |T(s)|$, computed on the original script and on SLP1
separately, since Devanagari is \numBytesPerDevanagariChar{} UTF-8 bytes per character and
SLP1 is one.

\paragraph{Parity.} For sentence-aligned corpora $A$ and $B$ of the same content, scored
under \emph{the same} tokenizer on both sides \citep{petrov-etal-2023-tokenizers}:
\begin{equation}
  \mathrm{Par}(T, A, B) = \frac{\sum_{i} |T(a_i)|}{\sum_{i} |T(b_i)|}.
\end{equation}

\paragraph{Tokens per proposition.} The headline. For aligned pairs $(s_i, e_i)$ of
Sanskrit source and English translation, scored under possibly \emph{different}
tokenizers $X$ and $Y$:
\begin{equation}
  \mathrm{TPP}(X, Y) = \frac{\sum_{i} |X(s_i)|}{\sum_{i} |Y(e_i)|}.
  \label{eq:tpp}
\end{equation}
A translated sentence pair is not a controlled propositional unit; it is the best
available proxy for one, and the translator's verbosity lands in the denominator. The
Limitations section returns to this.

\paragraph{The matched English control.} Equation \ref{eq:tpp} lets $Y$ be anything, and
the choice decides the answer. Setting $Y$ to a deployed English tokenizer measures
deployed practice, not language, since such a tokenizer differs from a Sanskrit arm in
vocabulary size and training domain as well as in language. We therefore train a control
family, \texttt{E1}, in which $Y$ shares the Sanskrit arm's algorithm, vocabulary size and
training corpus and differs only in which side of the parallel text it saw. Both readings
are reported; only the second supports a claim about Sanskrit.

\paragraph{Intervals.} Every tokens-per-proposition ratio carries a paired bootstrap
interval: resample the
aligned pairs with replacement, recompute the ratio of sums on each draw, and take the
\numCiLevel{}\% percentile interval over \numBootstrap{} draws, seeded at
\numBootstrapSeed{}. Pairing is what makes the interval meaningful, since the two sides of
a pair are the same content.
Sentences are not exchangeable within these corpora, so every controlled ratio carries a
second interval from a block bootstrap over non-overlapping blocks of \numBlockLength{}
consecutive pairs: Itih\=asa test is consecutive verses of one epic and FLORES devtest
consecutive sentences of the documents it was drawn from. The percentile interval stays
primary and is what every table prints; the block intervals are in Appendix
\ref{sec:blockci}, and the one reading that changes under them is named where it is made.
Counts of how many comparisons agree in sign are descriptive tallies and not tests.

We do not report perplexity anywhere. It is not comparable across tokenizers with
different vocabularies, and every comparison in this paper is across tokenizers.

\section{Experimental setup}
\label{sec:setup}

\paragraph{Corpora.} FLORES-200 devtest \citep{nllb-2022-flores} supplies
\numFloresSents{} sentences aligned across \texttt{san\_Deva}, \texttt{hin\_Deva} and
\texttt{eng\_Latn}. Its English side is the source and every other language a professional
translation of it, so it is the one corpus here in which both Indic sides stand in the same
relation to the text: the Sanskrit-against-Hindi comparison is on equal footing, while the
Sanskrit-against-English comparison still carries translation-length bias, in one
direction. S\=amayik
\citep{maheshwari-etal-2024-samayik} supplies contemporary English-Sanskrit prose:
\numSamayikTest{} test pairs and \numSamayikOod{} pairs in the split its release calls
\texttt{test\_ood}, out of domain with respect to the domains S\=amayik trains on, being
transcripts of the Mann ki Baat radio broadcasts. Itih\=asa
\citep{aralikatte-etal-2021-itihasa} supplies \numItihasaTest{} verse pairs. Prose is
primary and verse secondary, decided before any arm was run: meter constrains word choice
and inflates compounding, so verse cannot carry a claim about tokenization.

\begin{table}[t]
\centering
\footnotesize
\setlength{\tabcolsep}{3.2pt}
\begin{tabular}{lllll}
\toprule
Corpus & Source & Transl. & Inflates & Bias \\
\midrule
S\=amayik test & mixed & mixed & --- & unsigned \\
S\=amayik test\_ood & En & Sa & numer. & upward \\
Itih\=asa test & Sa & En & denom. & downward \\
FLORES devtest & En & Sa & numer. & upward \\
\bottomrule
\end{tabular}
\caption{Which side of each corpus is a translation, and the sign that puts on the
  tokens-per-proposition ratio. Translationese lengthens the translated side, so a
  translated English denominator biases the ratio downwards and a translated Sanskrit
  numerator upwards. S\=amayik test, the primary corpus for
  Section \ref{sec:rq2-controlled}, pools sub-corpora running in both directions and one,
  the New Testament portion, in neither, so its sign is unknown. On FLORES and on the
  out-of-domain split the bias runs towards the conclusion drawn there, and on Itih\=asa
  against the reading Section \ref{sec:verse} declines to make. These directions are read
  off each corpus's own documentation and are not measured here.}
\label{tab:direction}
\end{table}

The tokens-per-proposition denominator is therefore sometimes the source and sometimes
the translation, as Table \ref{tab:direction} records. Only S\=amayik test is mixed, and it
is mixed because its release states a direction per sub-corpus: Spoken Tutorials and Mann
ki Baat are English originals rendered into Sanskrit; G\={\i}t\=a Sop\=ana\.{m} was
translated into English in house; the New Testament portion pairs an English edition with
an 1851 Sanskrit translation, so neither side renders the other; and for the NIOS school
material no direction is stated at all. The out-of-domain split is not mixed, being Mann
ki Baat alone.

\paragraph{Arms.} \emph{Deployed general purpose:} \texttt{o200k\_base},
Llama-4, Gemma-3 and GPT-2. \emph{Deployed Indic:} Sarvam-1 \citep{sarvam-2024}, SUTRA
\citep{bendale-etal-2024-sutra} and BrahmicTokenizer-131K \citep{shravan-2026-brahmic},
the last of which is a single-author preprint plus a model card rather than a
peer-reviewed release, and is read here as a deployed artefact on that footing.
IndicSuperTokenizer is an arm this paper set out to include and is absent: none of the
three candidate repository identifiers resolves, so it is omitted from every table rather
than imputed. \emph{Trained here:} BPE \citep{gage-1994-algorithm,sennrich-etal-2016-neural}
and Unigram \citep{kudo-2018-subword,kudo-richardson-2018-sentencepiece} at
\numVocabSmall, \numVocabLarge{} and \numVocabHuge{} pieces, on the Sanskrit side of the
S\=amayik and Itih\=asa training splits. \emph{Matched control:} the same six recipes on
the English side of the same splits, and a second family of six on a subsample of that
English side cut to the Sanskrit corpus's byte count, which
Section \ref{sec:rq2-controlled} reads against the first.
\emph{Byte-level reference:} \texttt{T7\_byt5} takes UTF-8 bytes as its tokens,
\numTSevenVocab{} ids and no training at all. Scored on both sides of the same pairs it
reports the two sides' byte ratio rather than a comparison between tokenizers, which makes
it the no-vocabulary reference the framing of Claim B invites, and
Sections \ref{sec:rq2-controlled} and \ref{sec:verse} read it as one. Deployed vocabularies
range from \numVocabMin{} to
\numVocabMax{} ids, which is why they are reported as existing practice and never as a
controlled comparison. Appendix \ref{sec:arms} records what each arm loaded.

\paragraph{Substitutions to disclose.} The official Llama-4 and Gemma-3 repositories are
gated and the machine running these experiments has no access token, so both arms load
ungated re-uploads of the same releases; their id-space sizes match the published ones, but
the mirrors could not be byte-compared against the originals, because reading the originals
is what the gate prevents. FLORES-200 is read from the official NLLB tarball rather than
from a dataset hub, whose copies are gated as well.

\paragraph{Leakage control.} Every evaluation sentence, on \emph{both} sides, is hashed
into an exclusion list, and every training corpus is filtered against it before a
tokenizer sees a line. Each run re-verifies the check: of the \numCheckedSentences{}
evaluation pairs used here, \numLeakedSentences{} were missing from the list on either
side.

\paragraph{Reproducibility.} Every experiment is one command and writes a
\texttt{results.json} and its \texttt{config.yaml}, each recording the git commit and
whether the tree was clean; every table and figure below is regenerated from the tracked
snapshot by a script. The code and the snapshot are
\iffinalmode
at \url{https://github.com/DS436/sanskrit-token}.
\else
in an anonymised repository.
\fi

\section{Results}
\label{sec:results}

\subsection{RQ1: how large is the penalty?}
\label{sec:rq1}

Figure \ref{fig:parity} draws the parity ratios and Table \ref{tab:parity}, in
Appendix \ref{sec:fulltables}, tabulates them with the SLP1 column beside them.
Under the three deployed general-purpose tokenizers with vocabularies of
\numLargeVocabFloor{} ids or more,
Sanskrit costs \numParitySaEnLo--\numParitySaEnHi{} times the English tokens for the same
FLORES content. Under GPT-2, whose \numVocabMin-id vocabulary predates any serious
Devanagari coverage, it costs \numParitySaEnGptTwo{} times as many. Against Hindi, the
natural control since it shares the script but has no productive external sandhi and far
less inflectional fusion, though it does retain sandhi inside its Sanskrit loanwords,
the same tokenizers cost only \numParitySaHiLo--\numParitySaHiHi{} times as many tokens.

Two predictions from our design document fail here. Fertility above \numPreregFertLo{} on
Sanskrit is refuted for every arm with a vocabulary of \numLargeVocabFloor{} ids or more
(\numFertSaLo--\numFertSaHi, Table \ref{tab:fertcomp}, in Appendix \ref{sec:fulltables}
because fertility is reported here and never led with); it holds only for GPT-2 on raw
Devanagari (\numFertSaGptTwo), whose SLP1 variant falls back below \numPreregFertLo{}
(\numFertSaGptTwoSlpOne), so it is an old vocabulary meeting three-byte characters rather
than evidence about Sanskrit. A Sanskrit-over-Hindi parity above
\numPreregParityThreshold{} is refuted outright.

The gap between the two metrics is the substantive finding. Over these \numFloresSents{}
sentences Sanskrit is written in \numSaWords{} whitespace words against English's
\numEnWords{} and Hindi's \numHiWords, because sandhi and compounding fuse into one word
what the other two write as several. Dividing by that smaller denominator makes the
fertility-derived ratio overstate the penalty: Sanskrit over English reads
\numFertRatioSaEnLo--\numFertRatioSaEnHi{} by fertility against a measured parity of
\numParitySaEnLo--\numParitySaEnGptTwo, and Sanskrit over Hindi reads
\numFertRatioSaHiLo--\numFertRatioSaHiHi{} by fertility against a measured
\numParitySaHiLo--\numParitySaHiHi. Read off fertility, the Hindi hypothesis would have
been \emph{falsely confirmed} for all four arms. That is not a rounding difference; it is
the difference between counting tokens per word and counting tokens per unit of content.

Compression makes the same point from the byte side, on both encodings. Sanskrit in
Devanagari reads \numCompSaOrigLo--\numCompSaOrigHi{} bytes per token against English's
\numCompEnLo--\numCompEnHi, which compares nothing, Devanagari costing
\numBytesPerDevanagariChar{} UTF-8 bytes per character; in SLP1 the same text reads
\numCompSaSlpLo--\numCompSaSlpHi.

\begin{figure}[t]
  \centering
  \includegraphics[width=\columnwidth]{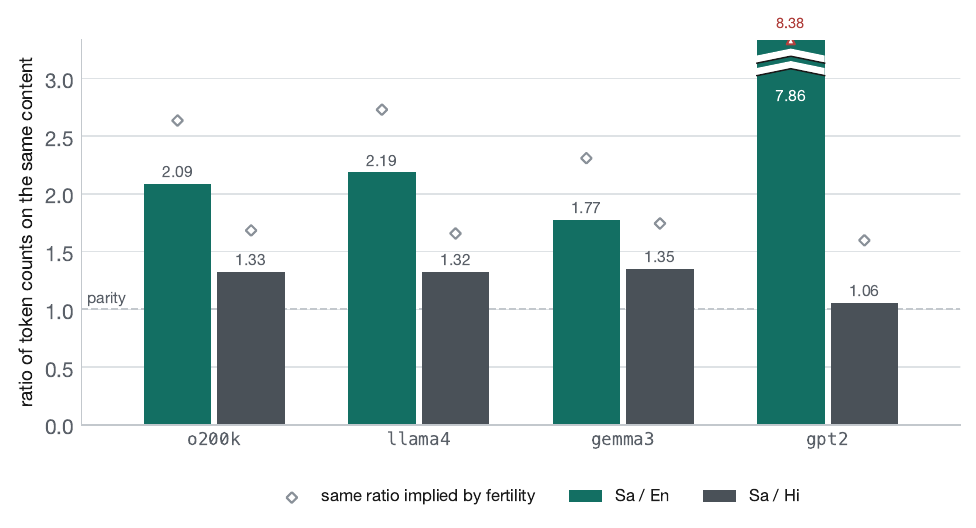}
  \caption{Measured parity (bars) against the ratio fertility implies (diamonds), on
    \numFloresSents{} aligned FLORES-200 devtest sentences, original script. Both series
    are ratios of token counts on the same content; the diamonds are what that ratio would
    have been had it been derived from fertility, and are not a cost per unit of content.
    Every diamond sits above its bar: fertility divides by a Sanskrit word count that
    sandhi and compounding make small, so it overstates the penalty. GPT-2's
    Sanskrit-over-English bar and diamond run off the axis and carry their values instead.}
  \label{fig:parity}
\end{figure}

\subsection{RQ2 under deployed practice}
\label{sec:rq2-deployed}

Every deployed arm, English-centric and Indic alike, puts Sanskrit above parity on prose
and on FLORES: across those corpora, with the Sanskrit side scored in SLP1 against the
deployed \texttt{o200k} pivot, the range is \numDeployedLo--\numDeployedHi{} with intervals
excluding 1.0 throughout, and on the primary prose corpus \texttt{o200k} against itself
reads \numDeployedSamayikOTwoHundredK{} \numDeployedSamayikOTwoHundredKCi{}
(Table \ref{tab:tppdeployed-samayik-test}, in Appendix \ref{sec:fulltables} with the other
corpora and the second pivot). This is the robust finding of the paper: under the
tokenizers currently shipped Sanskrit content costs that many times the English tokens for
the same propositions, and the Indic-specialised arms do not fix it.

The trained Sanskrit arms read very differently in the same table. BPE at
\numVocabLarge{} pieces comes in at \numTppBpeSixtyFourDeployed{}
\numTppBpeSixtyFourDeployedCi{}: below parity, with the interval clear of it, and the
\numVocabHuge-piece arm lower still. Taken at face value, that is the crossing our design
document predicted.

\begin{table*}[t]
\centering
\setlength{\tabcolsep}{3.4pt}
\scriptsize
\begin{tabular}{llllll}
\toprule
Matched pair & Control & S\=amayik test & S\=amayik test\_ood & Itih\=asa test & FLORES devtest \\
\midrule
BPE 32k & \texttt{E1} & 1.084 [1.070, 1.098] & 1.086 [1.072, 1.098] & \textbf{0.645 [0.642, 0.649]} & 1.143 [1.131, 1.154] \\
BPE 32k & \texttt{E1\_bm} & 1.081 [1.067, 1.096] & 1.085 [1.072, 1.097] & \textbf{0.645 [0.641, 0.648]} & 1.139 [1.127, 1.150] \\
BPE 64k & \texttt{E1} & 1.035 [1.021, 1.049] & 1.061 [1.048, 1.073] & \textbf{0.598 [0.594, 0.601]} & 1.144 [1.131, 1.155] \\
BPE 64k & \texttt{E1\_bm} & 1.030 [1.017, 1.043] & 1.055 [1.042, 1.066] & \textbf{0.597 [0.593, 0.600]} & 1.131 [1.119, 1.142] \\
BPE 128k & \texttt{E1} & \textbf{0.983 [0.971, 0.997]} & 1.025 [1.013, 1.037] & \textbf{0.557 [0.554, 0.560]} & 1.116 [1.103, 1.127] \\
BPE 128k & \texttt{E1\_bm} & \textbf{0.978 [0.965, 0.991]} & 1.015 [1.002, 1.026] & \textbf{0.555 [0.552, 0.559]} & 1.103 [1.091, 1.115] \\
Unigram 32k & \texttt{E1} & 1.142 [1.128, 1.157] & 1.163 [1.148, 1.176] & \textbf{0.658 [0.655, 0.661]} & 1.206 [1.194, 1.219] \\
Unigram 32k & \texttt{E1\_bm} & 1.133 [1.118, 1.148] & 1.159 [1.145, 1.173] & \textbf{0.651 [0.647, 0.654]} & 1.209 [1.196, 1.222] \\
Unigram 64k & \texttt{E1}$^{\ddagger}$ & 1.107 [1.092, 1.121] & 1.156 [1.142, 1.169] & \textbf{0.623 [0.619, 0.626]} & 1.219 [1.206, 1.232] \\
Unigram 64k & \texttt{E1\_bm}$^{\ddagger}$ & 1.088 [1.073, 1.102] & 1.138 [1.124, 1.150] & \textbf{0.613 [0.610, 0.617]} & 1.194 [1.181, 1.206] \\
Unigram 128k & \texttt{E1}$^{\ddagger}$ & 1.051 [1.037, 1.065] & 1.084 [1.071, 1.096] & \textbf{0.586 [0.584, 0.590]} & 1.145 [1.132, 1.157] \\
Unigram 128k & \texttt{E1\_bm}$^{\ddagger}$ & 1.033 [1.019, 1.047] & 1.067 [1.054, 1.079] & \textbf{0.578 [0.575, 0.581]} & 1.122 [1.110, 1.133] \\
\bottomrule
\end{tabular}
\caption{The controlled comparison. Each row is one matched pair: a trained Sanskrit arm over an English arm sharing its algorithm, vocabulary size and training corpus. \texttt{E1} is the pair-matched control, trained on the English side of the very sentences whose Sanskrit side trained the arm; \texttt{E1\_bm} is the byte-matched control, trained on a subsample of that text cut to the Sanskrit corpus's byte count. \texttt{BPE 32k} is \texttt{T1\_bpe\_raw\_32k} over \texttt{E1\_bpe\_32k}, and so on. Values are tokens per proposition with 95\% paired bootstrap intervals, Sanskrit scored in SLP1 and English as written; bold marks an interval entirely below 1.0. Pairs: S\=amayik test (prose) 2,417, S\=amayik test\_ood (prose, OOD) 4,047, Itih\=asa test (verse) 11,721, FLORES devtest 1,012. $^{\ddagger}$~marks a Unigram control the trainer could not bring to the requested size, settling at 50,659 and 62,896 pieces, which makes the English side dearer and pushes those rows down.}
\label{tab:tppcontrolled}
\end{table*}

\subsection{RQ2 under the matched control}
\label{sec:rq2-controlled}

It does not survive the control. Table \ref{tab:tppcontrolled} scores every trained
Sanskrit arm against its \texttt{E1} twins, and Figure \ref{fig:denominator} puts the
readings side by side. At \numVocabSmall{} and \numVocabLarge{} pieces, on S\=amayik test,
all \numControlledPairsSmall{} matched pairs, each size-matched arm against both controls,
sit above 1.0 with intervals excluding it,
\numTppControlledSamayikLo--\numTppControlledSamayikHi; the out-of-domain prose split gives
\numTppControlledOodLo--\numTppControlledOodHi{} and FLORES
\numTppControlledFloresLo--\numTppControlledFloresHi, in the same direction and with every
interval above 1.0. The \numVocabHuge-piece arms are taken separately below.

The Sanskrit numerator is the same token count in both ratios, by construction: the same
arm encodes the same sentences. The whole difference between \numTppBpeSixtyFourDeployed{}
and \numTppControlledBest{} is therefore the English denominator, and that much is
arithmetic rather than evidence; the measurement is the size of the change in it. The
\numVocabLarge-piece control arm spends \numEnTokensEOneBpe{} tokens on the English half of
S\=amayik test where the deployed tokenizer spends \numEnTokensOTwoHundredK, which is
\numEnTokenSavingPct\%{} fewer, because it too was trained on this domain. The apparent
Sanskrit advantage was the English pivot's handicap.

Matching on sentences leaves the control unmatched on everything else: the English side of
those sentences is \numBytesEnglishTrain{} MB against the Sanskrit side's
\numBytesSanskritTrain{} MB, \numBytesEnglishExcessPct\%{} more training text at the same
vocabulary size, so its cheaper tokenization could be read as the extra text rather than as
the language. The \texttt{E1\_bm} family closes that reading off: the same recipes on a
deterministic subsample of the same English corpus, shuffled at a fixed seed and cut to the
prefix whose written size first reaches the Sanskrit corpus's byte count, which is
\numBmLines{} lines, \numBmLinePct\%{} of them, \numBytesEnglishBmTrain{} MB. Every one of
the \numBmMoveCount{} ratios moves by at most \numBmMaxMove{} (median \numBmMedianMove),
\numBmDownwardCount{} of them downward, since a vocabulary trained on two thirds of the
corpus is slightly worse at it and a dearer denominator lowers the ratio; the number of
verdicts that change between the two controls is \numBmVerdictChanges. Neither family is
\emph{the} control, since matching sentences unmatches bytes and matching bytes unmatches
sentences, so both are reported for every pair and every reading here holds under either.

\begin{figure}[t]
  \centering
  \includegraphics[width=\columnwidth]{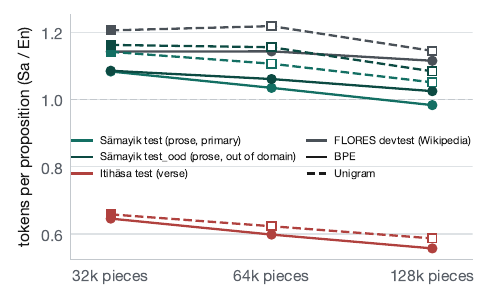}
  \caption{The controlled ratio against the vocabulary size it was measured at, per
    corpus, BPE solid and Unigram dashed, against the pair-matched control; each
    byte-matched twin sits within \numBmMaxMove{} of its point and is left out for
    legibility. Hollow markers are rows whose English control fell short of the requested
    size, which makes that side dearer and moves the point down. Only the size-matched BPE
    pair on in-domain prose crosses parity (dashed).}
  \label{fig:vocab}
\end{figure}

\begin{figure}[t]
  \centering
  \includegraphics[width=\columnwidth]{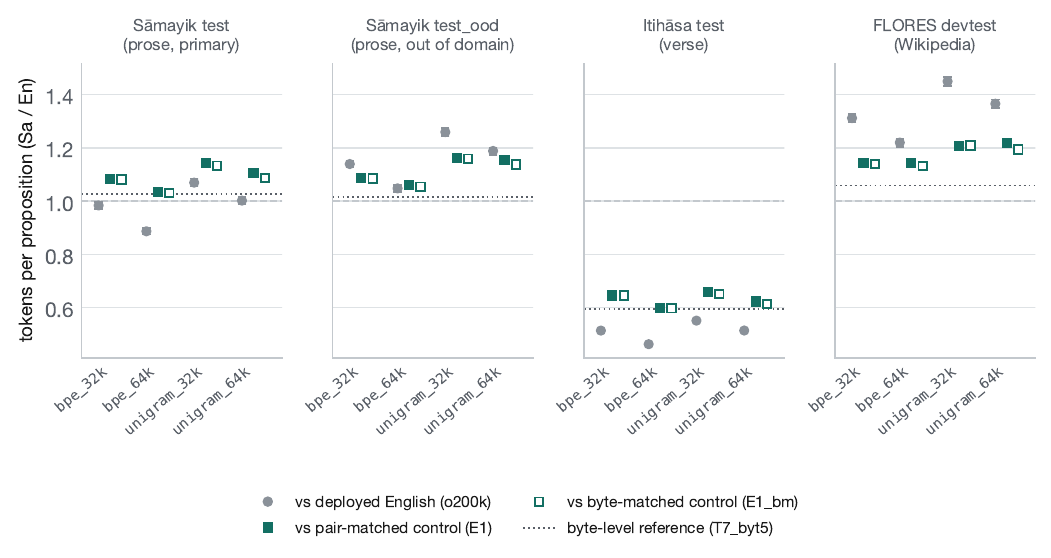}
  \caption{The same four Sanskrit arms, three English denominators, per corpus in the
    configured order (prose first). Circles score against the deployed English tokenizer,
    filled squares against the pair-matched control, hollow squares against the
    byte-matched one; the dashed line is parity, the dotted line the byte-level reference
    \texttt{T7\_byt5}, and \numCiLevel\%{} bootstrap intervals are narrower than the
    markers on most points. The in-domain prose panel moves from straddling parity to
    sitting above it when the denominator is controlled. Only the verse panel stays below,
    and there the pairs sit close to the byte reference rather than well above it. The
    \numVocabHuge-piece arms are in Figure \ref{fig:vocab}.}
  \label{fig:denominator}
\end{figure}

The vocabulary size is a different matter, and Figure \ref{fig:vocab} is why this section
names sizes rather than speaking of matched training in general. Raising the vocabulary
from \numVocabSmall{} to \numVocabHuge{} pieces lowers the controlled ratio on every corpus
and under both controls: the \numVocabHuge{} value is below both smaller ones in all
\numBpeSequencesTotal{} BPE sequences and falls at every step in
\numBpeMonotoneSequences{} of them, the exception being FLORES under the pair-matched
control, whose two smaller sizes differ by \numBpeFloresStepGap. On in-domain prose the
size-matched BPE pair crosses, at \numTppBpeHugeSamayik{} \numTppBpeHugeSamayikCi{}
pair-matched and \numTppBpeHugeSamayikBm{} \numTppBpeHugeSamayikBmCi{} byte-matched: the
first matched pairs here below 1.0 on prose with their intervals excluding it. The same
pair stays above parity out of domain, \numTppBpeHugeOod{} \numTppBpeHugeOodCi{} on the
out-of-domain prose split, whose block interval \numTppBpeHugeOodBlockCi{} does touch 1.0,
and \numTppBpeHugeFlores{} on FLORES, with the \numControlledPairsHuge{} pairs at that size
running \numTppControlledHugeOodLo--\numTppControlledHugeOodHi{} across the two corpora out
of domain for both sides. The controlled penalty therefore shrinks with vocabulary size,
and the claim this section supports is scoped to the sizes it was measured at:
matched-size raw subword training does not bring tokens per proposition below English on
prose at \numVocabSmall{} or \numVocabLarge{} pieces under either control, and at
\numVocabHuge{} it does so on in-domain prose alone.

The scope is narrower than the hypothesis it addresses. The only Sanskrit-native
tokenizers here are subword learners trained on raw, sandhied, compounded text with no
morphological information at all, which our design document lists as baselines rather than
as the arm it proposes; whether sandhi splitting or morpheme-constrained merges would cross
parity is untested here.

Two caveats keep even this provisional. The control equalises domain fit but not corpus
size or diversity, and neither side has seen a monolingual corpus. And the Unigram trainer
does not reach every size it is asked for: its English control settles at
\numUnigramSixtyFourPieces{} pieces at \numVocabLarge{} and again at \numVocabHuge, and at
\numUnigramBmPieces{} byte-matched, so those four rows are not size-matched and are marked
wherever they are printed. A smaller English vocabulary is dearer, so the shortfall pushes
those rows down rather than up, and they sit above 1.0 on prose regardless.

\begin{table*}[t]
\centering
\setlength{\tabcolsep}{4.0pt}
\scriptsize
\begin{tabular}{llrrrrrr}
\toprule
Matched pair & Control & S\=amayik test chars & density & TPP & Itih\=asa test chars & density & TPP \\
\midrule
BPE 32k & \texttt{E1} & 1.028 & 1.054 & 1.084 & 0.596 & 1.083 & 0.645 \\
BPE 32k & \texttt{E1\_bm} & 1.028 & 1.051 & 1.081 & 0.596 & 1.082 & 0.645 \\
BPE 64k & \texttt{E1} & 1.028 & 1.006 & 1.035 & 0.596 & 1.003 & 0.598 \\
BPE 64k & \texttt{E1\_bm} & 1.028 & 1.002 & 1.030 & 0.596 & 1.001 & 0.597 \\
Unigram 32k & \texttt{E1} & 1.028 & 1.111 & 1.142 & 0.596 & 1.104 & 0.658 \\
Unigram 32k & \texttt{E1\_bm} & 1.028 & 1.102 & 1.133 & 0.596 & 1.092 & 0.651 \\
Unigram 64k & \texttt{E1}$^{\ddagger}$ & 1.028 & 1.076 & 1.107 & 0.596 & 1.045 & 0.623 \\
Unigram 64k & \texttt{E1\_bm}$^{\ddagger}$ & 1.028 & 1.058 & 1.088 & 0.596 & 1.029 & 0.613 \\
BPE 128k & \texttt{E1} & 1.028 & 0.956 & 0.983 & 0.596 & 0.934 & 0.557 \\
\texttt{T7\_byt5} & --- & 1.028 & 0.999 & 1.027 & 0.596 & 0.998 & 0.595 \\
\bottomrule
\end{tabular}
\caption{Tokens per proposition factorised exactly, on the primary prose corpus and on the verse corpus. Per corpus: the character ratio $\sum_i c(s_i) / \sum_i c(e_i)$, how much text each side spends on the same propositions; the density ratio, each side's tokens per character over the other's; and their product, the ratio of Table~\ref{tab:tppcontrolled}. Sanskrit is read in SLP1, one character per phoneme, so the character ratio does not depend on the tokenizer and repeats down its column. \texttt{T7\_byt5} has no vocabulary to vary, so its product is the text ratio itself.}
\label{tab:decomposition}
\end{table*}

\subsection{Verse}
\label{sec:verse}

Itih\=asa is the one corpus where the crossing survives the control at every vocabulary
size: \numTppControlledItihasaAllLo--\numTppControlledItihasaAllHi{} over all
\numControlledPairsAll{} pairs, every interval excluding 1.0.

The ratio itself says where that comes from. Tokens per proposition factorises exactly into
the ratio of the two sides' character counts, how much text each side spends on the same
propositions, and the ratio of their tokens per character, how expensively each tokenizer
charges for a character of its own side; Table \ref{tab:decomposition} separates them. The
character ratio is \numCharRatioSamayik{} on the primary prose corpus and
\numCharRatioItihasa{} on Itih\=asa, a factor of \numCharRatioFactor{} apart, while the
density ratios barely move: \numDensityRatioSamayikLo--\numDensityRatioSamayikHi{} on prose
and \numDensityRatioItihasaLo--\numDensityRatioItihasaHi{} on verse at the two size-matched
vocabularies, agreeing pair for pair across the corpora to within \numDensityPairGap. The
byte-level reference says the same with no vocabulary at all: \numTSevenSamayik{} on prose
and \numTSevenItihasa{} on verse.

Three things follow. Matched subword tokenization charges the two languages nearly the same
per character, so a matched ratio mostly measures the relative length of the two texts. The
verse crossing therefore lives in the character ratio rather than in tokenization. And on
prose, Sanskrit in SLP1 is not shorter than English in characters
(\numCharRatioSamayik), so there is no character-level density there for a tokenizer to
recover; the in-domain crossing at \numVocabHuge{} pieces comes from the density factor
instead, where the Sanskrit side turns the cheaper per character
(\numDensityRatioSamayikHugeLo--\numDensityRatioSamayikHugeHi).

What the verse character ratio is remains open. Itih\=asa is \'sloka, and meter rewards
compounding, so its Sanskrit side may be short; its English side is a nineteenth-century
verse translation whose verbosity sits in the denominator, so it may be long. These corpora
cannot separate the two, and we do not attempt it: prose was designated primary before any
arm was run, and the verse number is a lead, not a finding.

\subsection{Tokens per proposition by sentence length}
\label{sec:length}

A corpus-level ratio compares whole corpora, and the two primary ones differ in length as
well as register: a S\=amayik test pair averages \numSamayikMeanEnWords{} English words to
\numSamayikMeanSaWords{} Sanskrit ones, while over Itih\=asa test the Sanskrit side is
barely longer, \numItihasaMeanSaWords{} words, against an English side of
\numItihasaMeanEnWords. That is the
character ratio of Table \ref{tab:decomposition} seen in words, and stratifying by length
probes bin by bin what that column states in one number per corpus. Binning is itself a
selection, so the strata are cut on each side in turn and read in pairs (Table
\ref{tab:tppbylength} and Figure \ref{fig:length}, in Appendix \ref{sec:lengthstrata}): a
high English word count selects pairs whose English side is long for their content, and
that side is the denominator, so the ratio falls as the bin rises whether or not density
changes with length, while binning on the Sanskrit side puts the same selection in the
numerator and makes it rise.

Every within-corpus gradient changes sign, \numLengthGradientsFlipped{} of
\numLengthGradientsTotal. On S\=amayik test the controlled ratio falls from
\numLengthEnFirstLo--\numLengthEnFirstHi{} at \numLengthEnFirstBin{} English words to
\numLengthEnLastLo--\numLengthEnLastHi{} at \numLengthEnLastBin, dropping below 1.0 for
\numLengthPairsBelowOneEn{} of the \numControlledPairs{} pairs, and rises from
\numLengthSaFirstLo--\numLengthSaFirstHi{} at \numLengthSaFirstBin{} Sanskrit words to
\numLengthSaLastLo--\numLengthSaLastHi{} at \numLengthSaLastBin{} with its
\numLengthPairsBelowOneSa{} sub-parity pairs now in the \emph{shortest} bin. We therefore
draw no claim about tokens per proposition changing with sentence length. Verse sits below
prose at every bin both corpora populate, under both stratifications
(\numVerseBelowProseComparisons{} of \numVerseBelowProseTotal{} comparisons, gaps
\numVerseProseGapMin--\numVerseProseGapMax), which is Section \ref{sec:verse}'s character
ratio surviving a control for length on either side rather than a second finding; a bin
equates the corpora only on the side it is cut on, so that separation stays bracketed
rather than isolated. Both counts are descriptive tallies and not tests, the matched pairs
being near duplicates that score the same Sanskrit sentences, and no multiplicity
correction is applied to them or to the intervals read against 1.0 elsewhere in this
paper.

Two diagnostics in the appendix support no claim here. R\'enyi efficiency
\citep{zouhar-etal-2023-tokenization} ranks the two trained BPE arms in the opposite order
from the controlled ratio and can be raised without improving the tokenization at all
\citep{cognetta-etal-2024-counterexamples}, so nothing rests on it (Appendix
\ref{sec:renyi}); Appendix \ref{sec:preregistration} sets our predictions against what was
measured, unedited.

\section{Conclusion}
\label{sec:conclusion}

Sanskrit's density is real at the word level and does not survive an English-centric
tokenizer. It does not come back under a matched English control either, at
\numVocabSmall{} or \numVocabLarge{} pieces and under either denominator; at
\numVocabHuge{} the BPE pair does read just below parity on in-domain prose and above it
elsewhere, so the negative result is scoped to the sizes measured rather than asserted of
raw subword training in general. Decomposing the ratio says how little there is here for
tokenization to win: matched tokenizers charge the two languages almost the same per
character, so a matched ratio mostly measures relative text length, and Sanskrit prose in
SLP1 is not shorter than its English translation. The follow-up tests what this leaves
open: sandhi splitting, and morpheme-constrained merges.

\section*{Limitations}

\paragraph{Verse and meter.} The one corpus on which Sanskrit stays below parity under the
control is verse, where meter is a confound that this design cannot separate from
tokenization. The length strata show that this number is not a sentence-length effect,
under binning on either side, but they do not identify what it is. We report the number and
decline to interpret it.

\paragraph{Translation length bias.} Tokens per proposition divides by a translation on one
side or the other, and a translation is systematically longer than its source
\citep{koppel-ordan-2011-translationese,volansky-etal-2015-features,%
graham-etal-2020-statistical}. The bias is therefore signed, and the signs are not all the
same (Table \ref{tab:direction}). On Itih\=asa the Sanskrit is the source and a
nineteenth-century English verse rendering is the denominator, so the bias runs against the
below-parity reading of Section \ref{sec:verse}, which we decline to make in any case. On
FLORES and on the out-of-domain prose split the English is the source and the Sanskrit the
translation, so the bias sits in the numerator and runs \emph{towards} the above-1.0
conclusion those rows support. On S\=amayik test, the primary corpus for that conclusion,
the direction is mixed by sub-corpus and the sign is unknown; nothing here bounds it, and
we do not claim it is small. FLORES does equalise the two Indic sides, since both are
translations of the same English source, so the Sanskrit-against-Hindi comparison there is
unaffected.

\paragraph{The Indic arms are deployed practice.} The Indic-specialised tokenizers differ
from every other arm in training corpus and vocabulary size. They answer what Sanskrit
costs under current Indic tooling. They cannot answer whether Indic-specialisation helps,
and we do not claim they do.

\paragraph{A single control language.} The control is English, one language, on the
English side of two corpora. Whether the same result holds against a matched Hindi
control, or against a matched control in a third morphologically rich language, is
untested.

\paragraph{No morphology claims.} Nothing here uses gold morpheme boundaries, so this
paper makes no claim about morphological alignment. Measuring whether a tokenizer's
boundaries agree with a language's own \citep{arnett-etal-2025-morphscore} requires a
gold-annotated corpus and is out of scope.

\paragraph{One corpus size.} Every trained arm saw the Sanskrit side, and every
pair-matched control arm the English side, of the same \numTrainPairsTotal{} training pairs
(S\=amayik \numSamayikTrain, Itih\=asa \numItihasaTrain), after exclusion filtering and
exact deduplication: \numBytesSanskritTrain{} MB of SLP1 against \numBytesEnglishTrain{} MB
of English, which is the asymmetry the byte-matched family, at
\numBytesEnglishBmTrain{} MB, exists to bracket. Tokenizer behaviour changes with corpus
scale, and none of these arms has been trained on a monolingual Sanskrit corpus of
realistic size. The corpus also sets a ceiling that two of the requested vocabulary sizes
ran into: the Unigram trainer stops at \numUnigramSixtyFourPieces{} pieces for the
pair-matched English control at both \numVocabLarge{} and \numVocabHuge, and at
\numUnigramBmPieces{} for the byte-matched one, so those four rows are not size-matched,
are marked wherever they are printed, and are biased downwards rather than upwards.

\iffinalmode
\section*{Acknowledgements}

We thank the creators and maintainers of the S\=amayik, Itih\=asa and FLORES-200 corpora
and of the tokenizers evaluated here for releasing them openly. The experimental pipeline,
analyses and manuscript were developed with substantial assistance from AI coding agents
(Claude, Anthropic), operating under the author's direction; all research decisions, the
interpretation of results and the final text are the author's responsibility, and every
number in the paper is regenerated by script from the public results snapshot.
\fi

\FloatBarrier

\bibliography{refs}

\appendix

\section{Full per-corpus tables}
\label{sec:fulltables}

\begin{table}[t]
\centering
\footnotesize
\begin{tabular}{lrrr}
\toprule
Arm & Sa/En & Sa/Hi & Sa (SLP1)/En \\
\midrule
\texttt{T0\_o200k} & 2.089 & 1.328 & 2.183 \\
\texttt{T0\_llama4} & 2.187 & 1.325 & 2.241 \\
\texttt{T0\_gemma3} & 1.774 & 1.353 & 2.189 \\
\texttt{T0\_gpt2} & 7.857 & 1.060 & 2.518 \\
\bottomrule
\end{tabular}
\caption{Parity ratios on 1,012 aligned FLORES-200 devtest sentences: Sanskrit tokens divided by the tokens the same tokenizer spends on the English or Hindi translation of the same sentence. Sa (SLP1)/En scores the SLP1 transliteration of the Sanskrit side against the same Latin-script English pivot. These four arms are deployed practice, not a controlled comparison: their vocabularies differ by more than five times (Appendix~\ref{sec:arms}).}
\label{tab:parity}
\end{table}

\begin{table*}[t]
\centering
\scriptsize
\begin{tabular}{llllll}
\toprule
Arm & Vocab. & SLP1 vs \texttt{o200k} & SLP1 vs Llama-4 & Orig.\ vs \texttt{o200k} & Fert. \\
\midrule
\texttt{T0\_o200k} & 200,019 & 1.835 [1.813, 1.858] & 1.802 [1.779, 1.824] & 1.901 [1.874, 1.926] & 3.15 \\
\texttt{T0\_llama4} & 201,135 & 1.882 [1.858, 1.905] & 1.848 [1.825, 1.871] & 1.985 [1.958, 2.011] & 3.21 \\
\texttt{T0\_gemma3} & 262,145 & 1.831 [1.809, 1.854] & 1.798 [1.776, 1.821] & 1.654 [1.632, 1.675] & 3.13 \\
\texttt{T0\_gpt2} & 50,257 & 2.123 [2.096, 2.151] & 2.085 [2.058, 2.113] & 6.562 [6.470, 6.652] & 3.57 \\
\texttt{T3\_sarvam} & 68,096 & 2.416 [2.386, 2.447] & 2.373 [2.342, 2.404] & 1.807 [1.782, 1.831] & 4.08 \\
\texttt{T3\_sutra} & 256,061 & 1.930 [1.905, 1.955] & 1.895 [1.870, 1.921] & 1.760 [1.735, 1.783] & 3.26 \\
\texttt{T3\_brahmic131k} & 131,072 & 1.873 [1.850, 1.896] & 1.839 [1.816, 1.862] & 1.899 [1.872, 1.924] & 3.22 \\
\texttt{T1\_bpe\_raw\_32k}$^{*}$ & 32,000 & \textbf{0.984 [0.970, 0.998]} & \textbf{0.966 [0.952, 0.980]} & n/a & 1.65 \\
\texttt{T1\_bpe\_raw\_64k}$^{*}$ & 64,000 & \textbf{0.887 [0.875, 0.899]} & \textbf{0.871 [0.859, 0.883]} & n/a & 1.49 \\
\texttt{T1\_bpe\_raw\_128k}$^{*}$ & 128,000 & \textbf{0.818 [0.807, 0.830]} & \textbf{0.804 [0.792, 0.815]} & n/a & 1.37 \\
\texttt{T2\_unigram\_raw\_32k}$^{*}$ & 32,000 & 1.069 [1.055, 1.085] & 1.050 [1.036, 1.066] & n/a & 1.80 \\
\texttt{T2\_unigram\_raw\_64k}$^{*}$ & 64,000 & 1.002 [0.988, 1.017] & \textbf{0.984 [0.970, 0.998]} & n/a & 1.69 \\
\texttt{T2\_unigram\_raw\_128k}$^{*}$ & 128,000 & \textbf{0.952 [0.939, 0.966]} & \textbf{0.935 [0.921, 0.949]} & n/a & 1.60 \\
\texttt{T7\_byt5} & 256 & 4.480 [4.424, 4.537] & 4.399 [4.344, 4.458] & 10.816 [10.672, 10.960] & 6.69 \\
\bottomrule
\end{tabular}
\caption{Deployed practice on S\=amayik test (prose), $n=2{,}417$ aligned pairs. Tokens per proposition: Sanskrit tokens under the named arm divided by English tokens under a deployed 200k-vocabulary English tokenizer, with 95\% paired bootstrap intervals. This is not a controlled comparison: vocabulary size and training domain vary alongside language, which is what Table~\ref{tab:tppcontrolled} holds fixed. $^{*}$~marks the provisional trained arms and \texttt{T7\_byt5} the byte-level reference. Fertility is in the last column for completeness and is not part of any claim here. Bold marks an interval entirely below 1.0. \texttt{T3\_indicsuper} is omitted: no candidate repository resolved.}
\label{tab:tppdeployed-samayik-test}
\end{table*}

\begin{table*}[t]
\centering
\scriptsize
\begin{tabular}{llllll}
\toprule
Arm & Vocab. & SLP1 vs \texttt{o200k} & SLP1 vs Llama-4 & Orig.\ vs \texttt{o200k} & Fert. \\
\midrule
\texttt{T0\_o200k} & 200,019 & 1.933 [1.911, 1.954] & 1.905 [1.883, 1.925] & 1.956 [1.933, 1.978] & 3.98 \\
\texttt{T0\_llama4} & 201,135 & 1.985 [1.962, 2.005] & 1.955 [1.933, 1.976] & 2.064 [2.039, 2.087] & 4.04 \\
\texttt{T0\_gemma3} & 262,145 & 1.949 [1.927, 1.970] & 1.920 [1.898, 1.942] & 1.676 [1.655, 1.693] & 3.98 \\
\texttt{T0\_gpt2} & 50,257 & 2.258 [2.231, 2.281] & 2.224 [2.199, 2.248] & 6.986 [6.903, 7.065] & 4.58 \\
\texttt{T3\_sarvam} & 68,096 & 2.556 [2.526, 2.585] & 2.518 [2.489, 2.547] & 1.798 [1.777, 1.818] & 5.17 \\
\texttt{T3\_sutra} & 256,061 & 2.057 [2.033, 2.081] & 2.027 [2.003, 2.049] & 1.827 [1.805, 1.847] & 4.16 \\
\texttt{T3\_brahmic131k} & 131,072 & 1.972 [1.949, 1.993] & 1.943 [1.921, 1.964] & 1.952 [1.929, 1.974] & 4.05 \\
\texttt{T1\_bpe\_raw\_32k}$^{*}$ & 32,000 & 1.139 [1.126, 1.152] & 1.123 [1.109, 1.134] & n/a & 2.31 \\
\texttt{T1\_bpe\_raw\_64k}$^{*}$ & 64,000 & 1.048 [1.036, 1.059] & 1.032 [1.020, 1.044] & n/a & 2.12 \\
\texttt{T1\_bpe\_raw\_128k}$^{*}$ & 128,000 & \textbf{0.977 [0.966, 0.988]} & \textbf{0.963 [0.951, 0.973]} & n/a & 1.98 \\
\texttt{T2\_unigram\_raw\_32k}$^{*}$ & 32,000 & 1.260 [1.244, 1.274] & 1.241 [1.226, 1.255] & n/a & 2.55 \\
\texttt{T2\_unigram\_raw\_64k}$^{*}$ & 64,000 & 1.189 [1.174, 1.202] & 1.171 [1.157, 1.184] & n/a & 2.40 \\
\texttt{T2\_unigram\_raw\_128k}$^{*}$ & 128,000 & 1.115 [1.101, 1.127] & 1.098 [1.085, 1.110] & n/a & 2.26 \\
\texttt{T7\_byt5} & 256 & 4.576 [4.522, 4.627] & 4.508 [4.455, 4.557] & 11.339 [11.206, 11.467] & 8.34 \\
\bottomrule
\end{tabular}
\caption{Deployed practice on S\=amayik test\_ood (prose, OOD), $n=4{,}047$ pairs. Columns as in Table~\ref{tab:tppdeployed-samayik-test}.}
\label{tab:tppdeployed-samayik-test-ood}
\end{table*}

\begin{table*}[t]
\centering
\scriptsize
\begin{tabular}{llllll}
\toprule
Arm & Vocab. & SLP1 vs \texttt{o200k} & SLP1 vs Llama-4 & Orig.\ vs \texttt{o200k} & Fert. \\
\midrule
\texttt{T0\_o200k} & 200,019 & 1.105 [1.100, 1.110] & 1.087 [1.082, 1.092] & 1.152 [1.147, 1.157] & 4.19 \\
\texttt{T0\_llama4} & 201,135 & 1.124 [1.119, 1.129] & 1.105 [1.100, 1.110] & 1.238 [1.232, 1.244] & 4.23 \\
\texttt{T0\_gemma3} & 262,145 & 1.095 [1.090, 1.100] & 1.077 [1.072, 1.082] & 0.996 [0.991, 1.000] & 4.12 \\
\texttt{T0\_gpt2} & 50,257 & 1.249 [1.243, 1.254] & 1.228 [1.223, 1.233] & 4.046 [4.027, 4.064] & 4.67 \\
\texttt{T3\_sarvam} & 68,096 & 1.425 [1.419, 1.432] & 1.402 [1.396, 1.408] & 1.116 [1.111, 1.121] & 5.28 \\
\texttt{T3\_sutra} & 256,061 & 1.169 [1.164, 1.174] & 1.150 [1.145, 1.155] & 1.092 [1.087, 1.097] & 4.33 \\
\texttt{T3\_brahmic131k} & 131,072 & 1.121 [1.115, 1.126] & 1.102 [1.097, 1.107] & 1.151 [1.146, 1.156] & 4.25 \\
\texttt{T1\_bpe\_raw\_32k}$^{*}$ & 32,000 & \textbf{0.513 [0.511, 0.516]} & \textbf{0.505 [0.502, 0.507]} & n/a & 1.90 \\
\texttt{T1\_bpe\_raw\_64k}$^{*}$ & 64,000 & \textbf{0.462 [0.460, 0.464]} & \textbf{0.454 [0.452, 0.457]} & n/a & 1.71 \\
\texttt{T1\_bpe\_raw\_128k}$^{*}$ & 128,000 & \textbf{0.425 [0.423, 0.427]} & \textbf{0.418 [0.416, 0.420]} & n/a & 1.57 \\
\texttt{T2\_unigram\_raw\_32k}$^{*}$ & 32,000 & \textbf{0.551 [0.548, 0.553]} & \textbf{0.542 [0.539, 0.544]} & n/a & 2.04 \\
\texttt{T2\_unigram\_raw\_64k}$^{*}$ & 64,000 & \textbf{0.513 [0.511, 0.516]} & \textbf{0.505 [0.502, 0.507]} & n/a & 1.90 \\
\texttt{T2\_unigram\_raw\_128k}$^{*}$ & 128,000 & \textbf{0.484 [0.481, 0.486]} & \textbf{0.476 [0.473, 0.478]} & n/a & 1.79 \\
\texttt{T7\_byt5} & 256 & 2.549 [2.538, 2.561] & 2.507 [2.496, 2.518] & 6.573 [6.542, 6.602] & 8.54 \\
\bottomrule
\end{tabular}
\caption{Deployed practice on Itih\=asa test (verse), $n=11{,}721$ pairs. Columns as in Table~\ref{tab:tppdeployed-samayik-test}.}
\label{tab:tppdeployed-itihasa-test}
\end{table*}

\begin{table*}[t]
\centering
\scriptsize
\begin{tabular}{llllll}
\toprule
Arm & Vocab. & SLP1 vs \texttt{o200k} & SLP1 vs Llama-4 & Orig.\ vs \texttt{o200k} & Fert. \\
\midrule
\texttt{T0\_o200k} & 200,019 & 2.183 [2.159, 2.207] & 2.169 [2.145, 2.192] & 2.089 [2.066, 2.112] & 3.50 \\
\texttt{T0\_llama4} & 201,135 & 2.256 [2.230, 2.280] & 2.241 [2.216, 2.265] & 2.201 [2.176, 2.225] & 3.56 \\
\texttt{T0\_gemma3} & 262,145 & 2.208 [2.184, 2.230] & 2.194 [2.170, 2.215] & 1.790 [1.769, 1.808] & 3.50 \\
\texttt{T0\_gpt2} & 50,257 & 2.534 [2.502, 2.562] & 2.518 [2.486, 2.544] & 7.907 [7.807, 7.999] & 3.97 \\
\texttt{T3\_sarvam} & 68,096 & 2.899 [2.865, 2.929] & 2.881 [2.847, 2.911] & 1.880 [1.858, 1.901] & 4.58 \\
\texttt{T3\_sutra} & 256,061 & 2.316 [2.288, 2.342] & 2.301 [2.273, 2.327] & 1.863 [1.840, 1.884] & 3.66 \\
\texttt{T3\_brahmic131k} & 131,072 & 2.244 [2.219, 2.269] & 2.230 [2.205, 2.254] & 2.087 [2.064, 2.110] & 3.57 \\
\texttt{T1\_bpe\_raw\_32k}$^{*}$ & 32,000 & 1.312 [1.297, 1.327] & 1.304 [1.288, 1.318] & n/a & 2.07 \\
\texttt{T1\_bpe\_raw\_64k}$^{*}$ & 64,000 & 1.220 [1.204, 1.234] & 1.212 [1.197, 1.226] & n/a & 1.92 \\
\texttt{T1\_bpe\_raw\_128k}$^{*}$ & 128,000 & 1.145 [1.130, 1.158] & 1.137 [1.123, 1.150] & n/a & 1.80 \\
\texttt{T2\_unigram\_raw\_32k}$^{*}$ & 32,000 & 1.450 [1.432, 1.467] & 1.441 [1.423, 1.458] & n/a & 2.29 \\
\texttt{T2\_unigram\_raw\_64k}$^{*}$ & 64,000 & 1.366 [1.349, 1.382] & 1.357 [1.341, 1.373] & n/a & 2.15 \\
\texttt{T2\_unigram\_raw\_128k}$^{*}$ & 128,000 & 1.283 [1.268, 1.298] & 1.275 [1.259, 1.290] & n/a & 2.02 \\
\texttt{T7\_byt5} & 256 & 5.203 [5.142, 5.259] & 5.170 [5.110, 5.226] & 12.906 [12.747, 13.053] & 7.29 \\
\bottomrule
\end{tabular}
\caption{Deployed practice on FLORES devtest, $n=1{,}012$ pairs. Columns as in Table~\ref{tab:tppdeployed-samayik-test}.}
\label{tab:tppdeployed-flores-devtest}
\end{table*}

\begin{table*}[t]
\centering
\footnotesize
\begin{tabular}{lll}
\toprule
Arm & Vocab. & Sa/Hi (original script) \\
\midrule
\texttt{T0\_o200k} & 200,019 & 1.328 [1.315, 1.342] \\
\texttt{T0\_llama4} & 201,135 & 1.325 [1.312, 1.339] \\
\texttt{T0\_gemma3} & 262,145 & 1.353 [1.339, 1.368] \\
\texttt{T0\_gpt2} & 50,257 & 1.060 [1.050, 1.071] \\
\texttt{T3\_sarvam} & 68,096 & 1.405 [1.390, 1.423] \\
\texttt{T3\_sutra} & 256,061 & 1.413 [1.398, 1.429] \\
\texttt{T3\_brahmic131k} & 131,072 & 1.327 [1.314, 1.341] \\
\bottomrule
\end{tabular}
\caption{Sanskrit over Hindi on FLORES devtest, both sides under the same tokenizer, in the original script. The corresponding SLP1 column is computed and stored in the results file but is not printed: SLP1 encodes the Sanskrit phoneme inventory, so Hindi characters outside it inflate the Hindi token count, which sits in this ratio's denominator and mechanically depresses it. Table~\ref{tab:fertcompfull} documents that failure and counts the sentences it affects. The trained arms are excluded by design: the Hindi pivot is defined over the deployed arms only.}
\label{tab:tpphindi}
\end{table*}

\begin{table*}[t]
\centering
\small
\begin{tabular}{lrrrrrrr}
\toprule
Arm & Sa & Hi & En & Sa/En & Sa/Hi & Sa bytes/tok. & En bytes/tok. \\
\midrule
\texttt{T0\_o200k} & 3.75 & 2.23 & 1.42 & 2.63 & 1.68 & 6.18 & 4.92 \\
\texttt{T0\_llama4} & 3.88 & 2.34 & 1.42 & 2.73 & 1.66 & 5.86 & 4.88 \\
\texttt{T0\_gemma3} & 3.11 & 1.79 & 1.35 & 2.31 & 1.74 & 7.21 & 4.87 \\
\texttt{T0\_gpt2} & 12.49 & 7.82 & 1.49 & 8.38 & 1.60 & 1.63 & 4.88 \\
\bottomrule
\end{tabular}
\caption{Fertility (tokens per whitespace word, columns 2--4) on the same sentences, the ratios that fertility implies (columns 5--6), and UTF-8 bytes per token (columns 7--8). Original script throughout. Fertility is reported for comparability with the literature and is not this paper's measure of cost: Sanskrit's denominator is its word count, which sandhi and compounding make small, so columns 5--6 overstate the penalty that Table~\ref{tab:parity} measures. Bytes per token are not comparable across scripts, since Devanagari is three UTF-8 bytes per character. The SLP1 variants of every column are in Table~\ref{tab:fertcompfull}.}
\label{tab:fertcomp}
\end{table*}

\begin{table*}[t]
\centering
\footnotesize
\begin{tabular}{lllrrrr}
\toprule
Arm & Language & Script & Fertility & s.d. & Bytes/tok. & Words \\
\midrule
\texttt{T0\_o200k} & \texttt{san\_Deva} & original & 3.75 & 1.97 & 6.18 & 16,975 \\
\texttt{T0\_o200k} & \texttt{san\_Deva} & slp1 & 3.50 & 1.92 & 2.38 & 16,975 \\
\texttt{T0\_o200k} & \texttt{hin\_Deva} & original & 2.23 & 1.15 & 7.97 & 25,643 \\
\texttt{T0\_o200k} & \texttt{hin\_Deva} & slp1$^{\dagger}$ & 2.56 & 1.29 & 2.31 & 25,643 \\
\texttt{T0\_o200k} & \texttt{eng\_Latn} & original & 1.42 & 0.72 & 4.92 & 21,901 \\
\texttt{T0\_llama4} & \texttt{san\_Deva} & original & 3.88 & 2.05 & 5.86 & 16,975 \\
\texttt{T0\_llama4} & \texttt{san\_Deva} & slp1 & 3.56 & 1.98 & 2.31 & 16,975 \\
\texttt{T0\_llama4} & \texttt{hin\_Deva} & original & 2.34 & 1.18 & 7.55 & 25,643 \\
\texttt{T0\_llama4} & \texttt{hin\_Deva} & slp1$^{\dagger}$ & 2.57 & 1.30 & 2.27 & 25,643 \\
\texttt{T0\_llama4} & \texttt{eng\_Latn} & original & 1.42 & 0.72 & 4.88 & 21,901 \\
\texttt{T0\_gemma3} & \texttt{san\_Deva} & original & 3.11 & 1.67 & 7.21 & 16,975 \\
\texttt{T0\_gemma3} & \texttt{san\_Deva} & slp1 & 3.50 & 1.94 & 2.36 & 16,975 \\
\texttt{T0\_gemma3} & \texttt{hin\_Deva} & original & 1.79 & 0.96 & 9.49 & 25,643 \\
\texttt{T0\_gemma3} & \texttt{hin\_Deva} & slp1$^{\dagger}$ & 2.52 & 1.30 & 2.30 & 25,643 \\
\texttt{T0\_gemma3} & \texttt{eng\_Latn} & original & 1.35 & 0.73 & 4.87 & 21,901 \\
\texttt{T0\_gpt2} & \texttt{san\_Deva} & original & 12.49 & 7.40 & 1.63 & 16,975 \\
\texttt{T0\_gpt2} & \texttt{san\_Deva} & slp1 & 3.97 & 2.23 & 2.05 & 16,975 \\
\texttt{T0\_gpt2} & \texttt{hin\_Deva} & original & 7.82 & 4.18 & 1.68 & 25,643 \\
\texttt{T0\_gpt2} & \texttt{hin\_Deva} & slp1$^{\dagger}$ & 2.85 & 1.49 & 2.02 & 25,643 \\
\texttt{T0\_gpt2} & \texttt{eng\_Latn} & original & 1.49 & 0.81 & 4.88 & 21,901 \\
\bottomrule
\end{tabular}
\caption{Fertility and compression for every arm, language and script variant. $^{\dagger}$~The Hindi SLP1 rows are approximate and should never be quoted as a measurement of Hindi: SLP1 encodes the Sanskrit phoneme inventory, so of the 1,012 Hindi sentences 513 contain a nukta consonant that the transliterator emits as a literal ASCII digit, and 256 of the resulting strings still contain unconverted Devanagari. Both effects inflate the Hindi token count. No parity number in this paper uses an SLP1 Hindi pivot.}
\label{tab:fertcompfull}
\end{table*}

\subsection{Length strata}
\label{sec:lengthstrata}

\begin{figure}[!tb]
  \centering
  \includegraphics[width=\columnwidth]{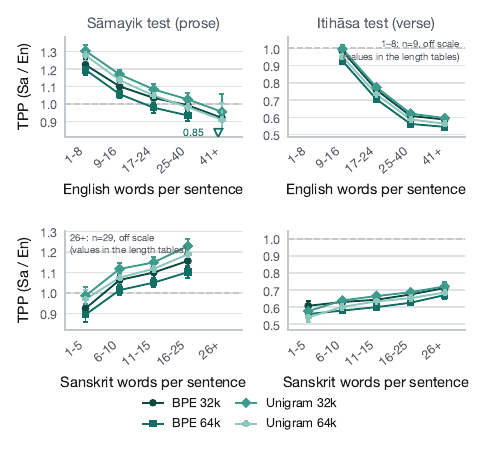}
  \caption{Tokens per proposition under the matched control by sentence length, prose left
    and verse right, binned on the English side's word count (top row) and on the
    Sanskrit side's (bottom). Marker shape and hue both identify the matched pair; markers
    are hollow where a bin holds fewer than \numLengthSparseBelow{} pairs, bars are
    \numCiLevel\%{} bootstrap intervals, and the dashed line is parity. Each panel is
    scaled to its own non-sparse bins, so positions are \emph{not} comparable across
    panels and that comparison is numeric, against Table \ref{tab:tppbylength}; a bin whose
    points all run off its panel carries a note instead of a stack of markers. Every
    gradient reverses between the rows, which is what selection on the binned side looks
    like and what a length effect is not.}
  \label{fig:length}
\end{figure}
\begin{table*}[t]
\centering
\setlength{\tabcolsep}{3.5pt}
\scriptsize
\begin{tabular}{lrrrrr}
\toprule
\multicolumn{6}{l}{\emph{(a) Bins cut on the English side}} \\
Matched pair & 1--8 En.\ words & 9--16 En.\ words & 17--24 En.\ words & 25--40 En.\ words & 41+ En.\ words \\
\midrule
\multicolumn{6}{l}{\emph{S\=amayik test (prose)}} \\
$n$ pairs & 835 & 995 & 389 & 187 & 11$^{\dagger}$ \\
BPE 32k & 1.225 [1.191, 1.260] & 1.101 [1.080, 1.122] & 1.035 [1.006, 1.064] & 0.992 [0.957, 1.025] & 0.920 [0.835, 1.006]$^{\dagger}$ \\
BPE 64k & 1.196 [1.165, 1.227] & 1.056 [1.036, 1.077] & 0.979 [0.950, 1.006] & 0.934 [0.900, 0.965] & 0.851 [0.762, 0.940]$^{\dagger}$ \\
Unigram 32k & 1.304 [1.269, 1.339] & 1.170 [1.146, 1.195] & 1.083 [1.052, 1.116] & 1.026 [0.986, 1.061] & 0.955 [0.853, 1.057]$^{\dagger}$ \\
Unigram 64k & 1.280 [1.248, 1.314] & 1.136 [1.113, 1.159] & 1.049 [1.016, 1.079] & 0.981 [0.941, 1.016] & 0.910 [0.818, 1.006]$^{\dagger}$ \\
\midrule
\multicolumn{6}{l}{\emph{Itih\=asa test (verse)}} \\
$n$ pairs & 9$^{\dagger}$ & 419 & 3,945 & 5,658 & 1,690 \\
BPE 32k & $^{\dagger}$ & 0.984 [0.964, 1.002] & 0.760 [0.755, 0.764] & 0.609 [0.605, 0.612] & 0.588 [0.579, 0.598] \\
BPE 64k & $^{\dagger}$ & 0.927 [0.908, 0.946] & 0.705 [0.701, 0.710] & 0.562 [0.559, 0.566] & 0.546 [0.536, 0.555] \\
Unigram 32k & $^{\dagger}$ & 0.996 [0.975, 1.016] & 0.774 [0.769, 0.779] & 0.622 [0.619, 0.626] & 0.596 [0.587, 0.605] \\
Unigram 64k & $^{\dagger}$ & 0.951 [0.933, 0.971] & 0.734 [0.729, 0.739] & 0.589 [0.585, 0.592] & 0.564 [0.555, 0.573] \\
\midrule
\multicolumn{6}{l}{\emph{(b) Bins cut on the Sanskrit side}} \\
Matched pair & 1--5 Sa.\ words & 6--10 Sa.\ words & 11--15 Sa.\ words & 16--25 Sa.\ words & 26+ Sa.\ words \\
\midrule
\multicolumn{6}{l}{\emph{S\=amayik test (prose)}} \\
$n$ pairs & 566 & 973 & 562 & 287 & 29$^{\dagger}$ \\
BPE 32k & 0.924 [0.885, 0.965] & 1.063 [1.039, 1.087] & 1.099 [1.074, 1.125] & 1.156 [1.128, 1.184] & 1.280 [1.196, 1.374]$^{\dagger}$ \\
BPE 64k & 0.894 [0.858, 0.934] & 1.013 [0.991, 1.037] & 1.050 [1.027, 1.074] & 1.101 [1.073, 1.130] & 1.206 [1.123, 1.304]$^{\dagger}$ \\
Unigram 32k & 0.986 [0.944, 1.030] & 1.117 [1.091, 1.144] & 1.148 [1.123, 1.176] & 1.228 [1.195, 1.263] & 1.371 [1.278, 1.492]$^{\dagger}$ \\
Unigram 64k & 0.968 [0.927, 1.013] & 1.075 [1.050, 1.101] & 1.117 [1.093, 1.144] & 1.188 [1.156, 1.222] & 1.319 [1.227, 1.433]$^{\dagger}$ \\
\midrule
\multicolumn{6}{l}{\emph{Itih\=asa test (verse)}} \\
$n$ pairs & 225 & 6,699 & 3,458 & 1,077 & 262 \\
BPE 32k & 0.608 [0.579, 0.638] & 0.630 [0.626, 0.634] & 0.643 [0.637, 0.649] & 0.674 [0.665, 0.685] & 0.712 [0.689, 0.736] \\
BPE 64k & 0.559 [0.534, 0.587] & 0.580 [0.576, 0.584] & 0.600 [0.594, 0.605] & 0.626 [0.617, 0.635] & 0.670 [0.646, 0.695] \\
Unigram 32k & 0.576 [0.549, 0.606] & 0.639 [0.635, 0.643] & 0.665 [0.659, 0.671] & 0.687 [0.678, 0.697] & 0.722 [0.700, 0.745] \\
Unigram 64k & 0.540 [0.514, 0.568] & 0.602 [0.598, 0.606] & 0.633 [0.627, 0.639] & 0.653 [0.643, 0.663] & 0.685 [0.664, 0.708] \\
\bottomrule
\end{tabular}
\caption{Tokens per proposition under the matched control, stratified by sentence length: (a) on the English side's whitespace word count (bin edges 1, 9, 17, 25, 41), (b) on the Sanskrit side's, in the original script (bin edges 1, 6, 11, 16, 26). Cells are the ratio with its 95\% paired bootstrap interval, and nothing is bolded, since which intervals clear 1.0 is read in the prose. Rows are the matched pairs of Table~\ref{tab:tppcontrolled} (BPE 32k is \texttt{T1\_bpe\_raw\_32k} over \texttt{E1\_bpe\_32k}); all four score the same sentences, so the $n$ row belongs to the bin. $^{\dagger}$~marks a bin under 30 pairs; under 10 it stands alone, the ratio left to Table~\ref{tab:tppbylengthall}. The halves are read together: binning on the English side selects pairs whose English side is long for their content, and that side is the denominator, so (a)'s gradient runs down whether or not density changes with length, while in (b) the binned side is the numerator and it runs up. The other two corpora are in Appendix~\ref{sec:lengthstrata}.}
\label{tab:tppbylength}
\end{table*}

\begin{table*}[t]
\centering
\setlength{\tabcolsep}{3.5pt}
\scriptsize
\begin{tabular}{lrrrrr}
\toprule
Matched pair & 1--8 En.\ words & 9--16 En.\ words & 17--24 En.\ words & 25--40 En.\ words & 41+ En.\ words \\
\midrule
\multicolumn{6}{l}{\emph{S\=amayik test (prose)}} \\
$n$ pairs & 835 & 995 & 389 & 187 & 11$^{\dagger}$ \\
BPE 32k & 1.225 [1.191, 1.260] & 1.101 [1.080, 1.122] & 1.035 [1.006, 1.064] & 0.992 [0.957, 1.025] & 0.920 [0.835, 1.006]$^{\dagger}$ \\
BPE 64k & 1.196 [1.165, 1.227] & 1.056 [1.036, 1.077] & 0.979 [0.950, 1.006] & 0.934 [0.900, 0.965] & 0.851 [0.762, 0.940]$^{\dagger}$ \\
Unigram 32k & 1.304 [1.269, 1.339] & 1.170 [1.146, 1.195] & 1.083 [1.052, 1.116] & 1.026 [0.986, 1.061] & 0.955 [0.853, 1.057]$^{\dagger}$ \\
Unigram 64k & 1.280 [1.248, 1.314] & 1.136 [1.113, 1.159] & 1.049 [1.016, 1.079] & 0.981 [0.941, 1.016] & 0.910 [0.818, 1.006]$^{\dagger}$ \\
\midrule
\multicolumn{6}{l}{\emph{S\=amayik test\_ood (prose, OOD)}} \\
$n$ pairs & 515 & 1,565 & 1,108 & 704 & 155 \\
BPE 32k & 1.245 [1.181, 1.313] & 1.142 [1.121, 1.164] & 1.103 [1.080, 1.125] & 1.036 [1.009, 1.059] & 0.951 [0.904, 0.995] \\
BPE 64k & 1.234 [1.172, 1.298] & 1.118 [1.097, 1.139] & 1.078 [1.054, 1.100] & 1.010 [0.984, 1.033] & 0.927 [0.882, 0.970] \\
Unigram 32k & 1.338 [1.272, 1.413] & 1.223 [1.199, 1.247] & 1.180 [1.155, 1.205] & 1.111 [1.083, 1.137] & 1.013 [0.960, 1.063] \\
Unigram 64k & 1.348 [1.283, 1.420] & 1.222 [1.199, 1.246] & 1.174 [1.149, 1.198] & 1.098 [1.069, 1.124] & 1.001 [0.950, 1.049] \\
\midrule
\multicolumn{6}{l}{\emph{Itih\=asa test (verse)}} \\
$n$ pairs & 9$^{\dagger}$ & 419 & 3,945 & 5,658 & 1,690 \\
BPE 32k & 8.944 [3.647, 21.501]$^{\dagger}$ & 0.984 [0.964, 1.002] & 0.760 [0.755, 0.764] & 0.609 [0.605, 0.612] & 0.588 [0.579, 0.598] \\
BPE 64k & 7.944 [3.061, 19.601]$^{\dagger}$ & 0.927 [0.908, 0.946] & 0.705 [0.701, 0.710] & 0.562 [0.559, 0.566] & 0.546 [0.536, 0.555] \\
Unigram 32k & 9.833 [4.104, 23.800]$^{\dagger}$ & 0.996 [0.975, 1.016] & 0.774 [0.769, 0.779] & 0.622 [0.619, 0.626] & 0.596 [0.587, 0.605] \\
Unigram 64k & 8.889 [3.350, 22.228]$^{\dagger}$ & 0.951 [0.933, 0.971] & 0.734 [0.729, 0.739] & 0.589 [0.585, 0.592] & 0.564 [0.555, 0.573] \\
\midrule
\multicolumn{6}{l}{\emph{FLORES devtest}} \\
$n$ pairs & 8$^{\dagger}$ & 244 & 464 & 280 & 16$^{\dagger}$ \\
BPE 32k & 1.165 [1.000, 1.336]$^{\dagger}$ & 1.155 [1.127, 1.181] & 1.148 [1.132, 1.165] & 1.136 [1.117, 1.155] & 1.091 [1.045, 1.139]$^{\dagger}$ \\
BPE 64k & 1.206 [1.047, 1.375]$^{\dagger}$ & 1.164 [1.136, 1.191] & 1.151 [1.133, 1.170] & 1.130 [1.111, 1.150] & 1.102 [1.057, 1.144]$^{\dagger}$ \\
Unigram 32k & 1.228 [1.086, 1.378]$^{\dagger}$ & 1.212 [1.181, 1.244] & 1.211 [1.191, 1.231] & 1.204 [1.183, 1.226] & 1.143 [1.100, 1.182]$^{\dagger}$ \\
Unigram 64k & 1.286 [1.104, 1.462]$^{\dagger}$ & 1.237 [1.209, 1.266] & 1.227 [1.209, 1.248] & 1.209 [1.188, 1.231] & 1.132 [1.082, 1.179]$^{\dagger}$ \\
\bottomrule
\end{tabular}
\caption{Tokens per proposition under the matched control, stratified by the \textbf{English} side's whitespace word count (bin edges 1, 9, 17, 25, 41). Each cell is the ratio with its 95\% paired bootstrap interval; nothing is bolded, since which intervals clear 1.0 is read in the prose. Rows are the matched pairs of Table~\ref{tab:tppcontrolled} in short form (BPE 32k is \texttt{T1\_bpe\_raw\_32k} over \texttt{E1\_bpe\_32k}, and so on), and all four score the same sentences, so the $n$ row belongs to the bin. $^{\dagger}$~marks a bin with fewer than 30 pairs. Every corpus, at the same bins as the corresponding half of the body's Table~\ref{tab:tppbylength}, which carries the two primary corpora. Read against its companion on the other side: a gradient that keeps its sign under both stratifications would be consistent with a length effect and one that changes sign is selection, and here all \numLengthGradientsTotal{} change sign.}
\label{tab:tppbylengthall}
\end{table*}

\begin{table*}[t]
\centering
\setlength{\tabcolsep}{3.5pt}
\scriptsize
\begin{tabular}{lrrrrr}
\toprule
Matched pair & 1--5 Sa.\ words & 6--10 Sa.\ words & 11--15 Sa.\ words & 16--25 Sa.\ words & 26+ Sa.\ words \\
\midrule
\multicolumn{6}{l}{\emph{S\=amayik test (prose)}} \\
$n$ pairs & 566 & 973 & 562 & 287 & 29$^{\dagger}$ \\
BPE 32k & 0.924 [0.885, 0.965] & 1.063 [1.039, 1.087] & 1.099 [1.074, 1.125] & 1.156 [1.128, 1.184] & 1.280 [1.196, 1.374]$^{\dagger}$ \\
BPE 64k & 0.894 [0.858, 0.934] & 1.013 [0.991, 1.037] & 1.050 [1.027, 1.074] & 1.101 [1.073, 1.130] & 1.206 [1.123, 1.304]$^{\dagger}$ \\
Unigram 32k & 0.986 [0.944, 1.030] & 1.117 [1.091, 1.144] & 1.148 [1.123, 1.176] & 1.228 [1.195, 1.263] & 1.371 [1.278, 1.492]$^{\dagger}$ \\
Unigram 64k & 0.968 [0.927, 1.013] & 1.075 [1.050, 1.101] & 1.117 [1.093, 1.144] & 1.188 [1.156, 1.222] & 1.319 [1.227, 1.433]$^{\dagger}$ \\
\midrule
\multicolumn{6}{l}{\emph{S\=amayik test\_ood (prose, OOD)}} \\
$n$ pairs & 571 & 1,500 & 1,046 & 738 & 192 \\
BPE 32k & 0.769 [0.727, 0.816] & 0.998 [0.979, 1.018] & 1.101 [1.081, 1.122] & 1.178 [1.151, 1.206] & 1.268 [1.212, 1.326] \\
BPE 64k & 0.759 [0.717, 0.806] & 0.978 [0.959, 0.997] & 1.076 [1.056, 1.097] & 1.147 [1.122, 1.174] & 1.238 [1.183, 1.295] \\
Unigram 32k & 0.820 [0.775, 0.870] & 1.067 [1.044, 1.087] & 1.183 [1.160, 1.208] & 1.260 [1.231, 1.291] & 1.361 [1.301, 1.427] \\
Unigram 64k & 0.815 [0.770, 0.866] & 1.065 [1.045, 1.085] & 1.174 [1.153, 1.198] & 1.249 [1.222, 1.279] & 1.349 [1.290, 1.413] \\
\midrule
\multicolumn{6}{l}{\emph{Itih\=asa test (verse)}} \\
$n$ pairs & 225 & 6,699 & 3,458 & 1,077 & 262 \\
BPE 32k & 0.608 [0.579, 0.638] & 0.630 [0.626, 0.634] & 0.643 [0.637, 0.649] & 0.674 [0.665, 0.685] & 0.712 [0.689, 0.736] \\
BPE 64k & 0.559 [0.534, 0.587] & 0.580 [0.576, 0.584] & 0.600 [0.594, 0.605] & 0.626 [0.617, 0.635] & 0.670 [0.646, 0.695] \\
Unigram 32k & 0.576 [0.549, 0.606] & 0.639 [0.635, 0.643] & 0.665 [0.659, 0.671] & 0.687 [0.678, 0.697] & 0.722 [0.700, 0.745] \\
Unigram 64k & 0.540 [0.514, 0.568] & 0.602 [0.598, 0.606] & 0.633 [0.627, 0.639] & 0.653 [0.643, 0.663] & 0.685 [0.664, 0.708] \\
\midrule
\multicolumn{6}{l}{\emph{FLORES devtest}} \\
$n$ pairs & 5$^{\dagger}$ & 153 & 311 & 457 & 86 \\
BPE 32k & 1.000 [0.868, 1.228]$^{\dagger}$ & 1.073 [1.043, 1.105] & 1.101 [1.082, 1.122] & 1.157 [1.143, 1.172] & 1.223 [1.182, 1.269] \\
BPE 64k & 0.967 [0.864, 1.119]$^{\dagger}$ & 1.075 [1.044, 1.109] & 1.098 [1.078, 1.119] & 1.160 [1.144, 1.175] & 1.225 [1.185, 1.267] \\
Unigram 32k & 1.138 [0.875, 1.398]$^{\dagger}$ & 1.132 [1.098, 1.167] & 1.164 [1.142, 1.188] & 1.221 [1.203, 1.238] & 1.287 [1.242, 1.334] \\
Unigram 64k & 1.095 [0.903, 1.275]$^{\dagger}$ & 1.148 [1.112, 1.187] & 1.172 [1.150, 1.196] & 1.237 [1.218, 1.253] & 1.295 [1.248, 1.346] \\
\bottomrule
\end{tabular}
\caption{Tokens per proposition under the matched control, stratified by the \textbf{Sanskrit} side's whitespace word count in the original script (bin edges 1, 6, 11, 16, 26). Cells, rows and daggers are as in Table~\ref{tab:tppbylength}. Every corpus, at the same bins as the corresponding half of the body's Table~\ref{tab:tppbylength}, which carries the two primary corpora. Read against its companion on the other side: a gradient that keeps its sign under both stratifications would be consistent with a length effect and one that changes sign is selection, and here all \numLengthGradientsTotal{} change sign.}
\label{tab:tppbylengthallsa}
\end{table*}

\subsection{Block-resampled intervals}
\label{sec:blockci}

Table \ref{tab:blockci} puts every controlled ratio's block interval beside its i.i.d.\ one.
Blocking widens the intervals on three of the four corpora, by
\numBlockWidenItihasaLo--\numBlockWidenItihasaHi$\times$ on Itih\=asa,
\numBlockWidenFloresLo--\numBlockWidenFloresHi$\times$ on FLORES and
\numBlockWidenOodLo--\numBlockWidenOodHi$\times$ on the out-of-domain prose split, which is
what local dependence between neighbouring sentences looks like. On S\=amayik test it does
not: the ratio of widths is \numBlockWidenSamayikLo--\numBlockWidenSamayikHi$\times$, and
\numBlockNarrowerSamayik{} of the \numBlockRowsPerCorpus{} rows are marginally narrower
under blocks than under pairs, which is noise in the estimate of the variance at that many
blocks rather than evidence of less of it, and is reported rather than hidden. Every verse
verdict survives, the highest block upper bound being \numBlockItihasaMaxUpper. One reading
changes, and Section \ref{sec:rq2-controlled} states it where it is made: the BPE pair at
\numVocabHuge{} pieces on the out-of-domain prose split is \numTppBpeHugeOod{}
\numTppBpeHugeOodCi{} under pairs and \numTppBpeHugeOodBlockCi{} under blocks, so it touches
parity there. Its byte-matched twin and the byte-level reference do the same on that split,
and no other interval changes the side of 1.0 it sits on.

\begin{table*}[t]
\centering
\setlength{\tabcolsep}{4pt}
\scriptsize
\begin{tabular}{llrrrr}
\toprule
Matched pair & Control & TPP & i.i.d.\ interval & Block interval & Width \\
\midrule
\multicolumn{6}{l}{\emph{S\=amayik test (prose)}, 2,417 pairs, 49 blocks} \\
BPE 32k & \texttt{E1} & 1.084 & [1.070, 1.098] & [1.070, 1.097] & 0.96$\times$ \\
BPE 32k & \texttt{E1\_bm} & 1.081 & [1.067, 1.096] & [1.067, 1.095] & 0.98$\times$ \\
BPE 64k & \texttt{E1} & 1.035 & [1.021, 1.049] & [1.022, 1.049] & 0.97$\times$ \\
BPE 64k & \texttt{E1\_bm} & 1.030 & [1.017, 1.043] & [1.018, 1.044] & 1.00$\times$ \\
BPE 128k & \texttt{E1} & 0.983 & [0.971, 0.997] & [0.971, 0.997] & 1.01$\times$ \\
BPE 128k & \texttt{E1\_bm} & 0.978 & [0.965, 0.991] & [0.965, 0.991] & 0.98$\times$ \\
Unigram 32k & \texttt{E1} & 1.142 & [1.128, 1.157] & [1.129, 1.156] & 0.95$\times$ \\
Unigram 32k & \texttt{E1\_bm} & 1.133 & [1.118, 1.148] & [1.120, 1.147] & 0.90$\times$ \\
Unigram 64k & \texttt{E1}$^{\ddagger}$ & 1.107 & [1.092, 1.121] & [1.093, 1.121] & 0.98$\times$ \\
Unigram 64k & \texttt{E1\_bm}$^{\ddagger}$ & 1.088 & [1.073, 1.102] & [1.075, 1.102] & 0.97$\times$ \\
Unigram 128k & \texttt{E1}$^{\ddagger}$ & 1.051 & [1.037, 1.065] & [1.037, 1.065] & 1.00$\times$ \\
Unigram 128k & \texttt{E1\_bm}$^{\ddagger}$ & 1.033 & [1.019, 1.047] & [1.020, 1.047] & 0.95$\times$ \\
\texttt{T7\_byt5} & --- & 1.027 & [1.016, 1.039] & [1.015, 1.037] & 0.98$\times$ \\
\midrule
\multicolumn{6}{l}{\emph{S\=amayik test\_ood (prose, OOD)}, 4,047 pairs, 81 blocks} \\
BPE 32k & \texttt{E1} & 1.086 & [1.072, 1.098] & [1.059, 1.113] & 2.12$\times$ \\
BPE 32k & \texttt{E1\_bm} & 1.085 & [1.072, 1.097] & [1.058, 1.112] & 2.15$\times$ \\
BPE 64k & \texttt{E1} & 1.061 & [1.048, 1.073] & [1.035, 1.088] & 2.17$\times$ \\
BPE 64k & \texttt{E1\_bm} & 1.055 & [1.042, 1.066] & [1.028, 1.081] & 2.16$\times$ \\
BPE 128k & \texttt{E1} & 1.025 & [1.013, 1.037] & [0.999, 1.050] & 2.13$\times$ \\
BPE 128k & \texttt{E1\_bm} & 1.015 & [1.002, 1.026] & [0.989, 1.039] & 2.13$\times$ \\
Unigram 32k & \texttt{E1} & 1.163 & [1.148, 1.176] & [1.130, 1.195] & 2.30$\times$ \\
Unigram 32k & \texttt{E1\_bm} & 1.159 & [1.145, 1.173] & [1.128, 1.191] & 2.29$\times$ \\
Unigram 64k & \texttt{E1}$^{\ddagger}$ & 1.156 & [1.142, 1.169] & [1.124, 1.187] & 2.35$\times$ \\
Unigram 64k & \texttt{E1\_bm}$^{\ddagger}$ & 1.138 & [1.124, 1.150] & [1.106, 1.169] & 2.37$\times$ \\
Unigram 128k & \texttt{E1}$^{\ddagger}$ & 1.084 & [1.071, 1.096] & [1.054, 1.114] & 2.35$\times$ \\
Unigram 128k & \texttt{E1\_bm}$^{\ddagger}$ & 1.067 & [1.054, 1.079] & [1.037, 1.097] & 2.37$\times$ \\
\texttt{T7\_byt5} & --- & 1.016 & [1.005, 1.027] & [0.996, 1.037] & 1.79$\times$ \\
\midrule
\multicolumn{6}{l}{\emph{Itih\=asa test (verse)}, 11,721 pairs, 235 blocks} \\
BPE 32k & \texttt{E1} & 0.645 & [0.642, 0.649] & [0.637, 0.654] & 2.47$\times$ \\
BPE 32k & \texttt{E1\_bm} & 0.645 & [0.641, 0.648] & [0.637, 0.653] & 2.45$\times$ \\
BPE 64k & \texttt{E1} & 0.598 & [0.594, 0.601] & [0.590, 0.606] & 2.40$\times$ \\
BPE 64k & \texttt{E1\_bm} & 0.597 & [0.593, 0.600] & [0.589, 0.605] & 2.48$\times$ \\
BPE 128k & \texttt{E1} & 0.557 & [0.554, 0.560] & [0.549, 0.564] & 2.43$\times$ \\
BPE 128k & \texttt{E1\_bm} & 0.555 & [0.552, 0.559] & [0.548, 0.563] & 2.41$\times$ \\
Unigram 32k & \texttt{E1} & 0.658 & [0.655, 0.661] & [0.650, 0.667] & 2.50$\times$ \\
Unigram 32k & \texttt{E1\_bm} & 0.651 & [0.647, 0.654] & [0.643, 0.660] & 2.52$\times$ \\
Unigram 64k & \texttt{E1}$^{\ddagger}$ & 0.623 & [0.619, 0.626] & [0.615, 0.631] & 2.45$\times$ \\
Unigram 64k & \texttt{E1\_bm}$^{\ddagger}$ & 0.613 & [0.610, 0.617] & [0.606, 0.622] & 2.45$\times$ \\
Unigram 128k & \texttt{E1}$^{\ddagger}$ & 0.586 & [0.584, 0.590] & [0.579, 0.595] & 2.47$\times$ \\
Unigram 128k & \texttt{E1\_bm}$^{\ddagger}$ & 0.578 & [0.575, 0.581] & [0.571, 0.586] & 2.47$\times$ \\
\texttt{T7\_byt5} & --- & 0.595 & [0.592, 0.598] & [0.588, 0.603] & 2.59$\times$ \\
\midrule
\multicolumn{6}{l}{\emph{FLORES devtest}, 1,012 pairs, 21 blocks} \\
BPE 32k & \texttt{E1} & 1.143 & [1.131, 1.154] & [1.113, 1.172] & 2.54$\times$ \\
BPE 32k & \texttt{E1\_bm} & 1.139 & [1.127, 1.150] & [1.109, 1.168] & 2.64$\times$ \\
BPE 64k & \texttt{E1} & 1.144 & [1.131, 1.155] & [1.114, 1.173] & 2.51$\times$ \\
BPE 64k & \texttt{E1\_bm} & 1.131 & [1.119, 1.142] & [1.101, 1.162] & 2.61$\times$ \\
BPE 128k & \texttt{E1} & 1.116 & [1.103, 1.127] & [1.086, 1.146] & 2.50$\times$ \\
BPE 128k & \texttt{E1\_bm} & 1.103 & [1.091, 1.115] & [1.074, 1.133] & 2.52$\times$ \\
Unigram 32k & \texttt{E1} & 1.206 & [1.194, 1.219] & [1.176, 1.235] & 2.33$\times$ \\
Unigram 32k & \texttt{E1\_bm} & 1.209 & [1.196, 1.222] & [1.177, 1.239] & 2.42$\times$ \\
Unigram 64k & \texttt{E1}$^{\ddagger}$ & 1.219 & [1.206, 1.232] & [1.189, 1.250] & 2.39$\times$ \\
Unigram 64k & \texttt{E1\_bm}$^{\ddagger}$ & 1.194 & [1.181, 1.206] & [1.164, 1.223] & 2.33$\times$ \\
Unigram 128k & \texttt{E1}$^{\ddagger}$ & 1.145 & [1.132, 1.157] & [1.117, 1.174] & 2.31$\times$ \\
Unigram 128k & \texttt{E1\_bm}$^{\ddagger}$ & 1.122 & [1.110, 1.133] & [1.094, 1.150] & 2.47$\times$ \\
\texttt{T7\_byt5} & --- & 1.059 & [1.049, 1.067] & [1.039, 1.078] & 2.11$\times$ \\
\bottomrule
\end{tabular}
\caption{Every controlled ratio under both resampling schemes. The 95\% i.i.d.\ interval resamples aligned pairs, and is the interval every other table in this paper reports; the block interval resamples non-overlapping blocks of 50 consecutive pairs in corpus order, the final short block kept, and is reported because these corpora are not exchangeable sentence by sentence: Itih\=asa test is consecutive verses of one epic and FLORES devtest consecutive sentences of the documents it was drawn from. Width is the block interval's width over the i.i.d.\ one. Rows, controls and the $^{\ddagger}$ mark are as in Table~\ref{tab:tppcontrolled}, with \texttt{T7\_byt5}, the byte-level reference, added at the foot of each block.}
\label{tab:blockci}
\end{table*}

\section{Arm details}
\label{sec:arms}

\begin{table*}[t]
\centering
\footnotesize
\begin{tabular}{llrl}
\toprule
Arm & Family & Vocab. & What loaded \\
\midrule
\texttt{E1\_bpe\_128k\_bm} & E1 & 128,000 & trained here; \texttt{tokenizer.json} sha256 \texttt{7efa70c72f69} \\
\texttt{E1\_bpe\_128k} & E1 & 128,000 & trained here; \texttt{tokenizer.json} sha256 \texttt{910c7dbf5185} \\
\texttt{E1\_bpe\_32k\_bm} & E1 & 32,000 & trained here; \texttt{tokenizer.json} sha256 \texttt{913ef268d3c4} \\
\texttt{E1\_bpe\_32k} & E1 & 32,000 & trained here; \texttt{tokenizer.json} sha256 \texttt{f639ee791e50} \\
\texttt{E1\_bpe\_64k\_bm} & E1 & 64,000 & trained here; \texttt{tokenizer.json} sha256 \texttt{eca779d73352} \\
\texttt{E1\_bpe\_64k} & E1 & 64,000 & trained here; \texttt{tokenizer.json} sha256 \texttt{e57168ce7f9d} \\
\texttt{E1\_unigram\_128k\_bm} & E1 & 50,659 & trained here; \texttt{tokenizer.json} sha256 \texttt{34fe0d224522} \\
\texttt{E1\_unigram\_128k} & E1 & 62,896 & trained here; \texttt{tokenizer.json} sha256 \texttt{b4638805204b} \\
\texttt{E1\_unigram\_32k\_bm} & E1 & 32,000 & trained here; \texttt{tokenizer.json} sha256 \texttt{c17367406798} \\
\texttt{E1\_unigram\_32k} & E1 & 32,000 & trained here; \texttt{tokenizer.json} sha256 \texttt{62f737185c0c} \\
\texttt{E1\_unigram\_64k\_bm} & E1 & 50,659 & trained here; \texttt{tokenizer.json} sha256 \texttt{1da233ae99c1} \\
\texttt{E1\_unigram\_64k} & E1 & 62,896 & trained here; \texttt{tokenizer.json} sha256 \texttt{b21ecf863056} \\
\texttt{T0\_gemma3} & T0 & 262,145 & \texttt{unsloth/gemma-3-4b-it} \\
\texttt{T0\_gpt2} & T0 & 50,257 & \texttt{openai-community/gpt2} \\
\texttt{T0\_llama4} & T0 & 201,135 & \texttt{unsloth/Llama-4-Scout-17B-16E-Instruct} \\
\texttt{T0\_o200k} & T0 & 200,019 & \texttt{o200k\_base} \\
\texttt{T1\_bpe\_raw\_128k}$^{*}$ & T1 & 128,000 & trained here; \texttt{tokenizer.json} sha256 \texttt{59ab04d3630f} \\
\texttt{T1\_bpe\_raw\_32k}$^{*}$ & T1 & 32,000 & trained here; \texttt{tokenizer.json} sha256 \texttt{e9ae328e26db} \\
\texttt{T1\_bpe\_raw\_64k}$^{*}$ & T1 & 64,000 & trained here; \texttt{tokenizer.json} sha256 \texttt{f74207d923a7} \\
\texttt{T2\_unigram\_raw\_128k}$^{*}$ & T2 & 128,000 & trained here; \texttt{tokenizer.json} sha256 \texttt{9ea97f4ea05c} \\
\texttt{T2\_unigram\_raw\_32k}$^{*}$ & T2 & 32,000 & trained here; \texttt{tokenizer.json} sha256 \texttt{09bc885de2ea} \\
\texttt{T2\_unigram\_raw\_64k}$^{*}$ & T2 & 64,000 & trained here; \texttt{tokenizer.json} sha256 \texttt{199e32f30ec1} \\
\texttt{T3\_brahmic131k} & T3 & 131,072 & \texttt{theschoolofai/BrahmicTokenizer-131K} \\
\texttt{T3\_sarvam} & T3 & 68,096 & \texttt{sarvamai/sarvam-1} \\
\texttt{T3\_sutra} & T3 & 256,061 & \texttt{TWO/sutra-mlt256-v2} \\
\texttt{T7\_byt5} & T7 & 256 & \texttt{bytes/utf-8} \\
\bottomrule
\end{tabular}
\caption{Every arm, the identifier that actually loaded, and its id-space size ($\mathrm{len}(\mathrm{tokenizer})$, which counts added special tokens). \texttt{T0\_llama4} and \texttt{T0\_gemma3} load ungated re-uploads of the official releases because the official repositories are gated and the machine running these experiments has no access token; the vocabulary sizes match the published ones, and the mirrors could not be byte-compared against the originals because reading the originals is what the gate prevents. A Hub repository is mutable and these rows carry no revision hash, so the identifier names what was loaded and not which bytes: the load date is the \texttt{timestamp} field of the results file the table is generated from. The sha256 discipline applies only to the arms trained here, whose \texttt{tokenizer.json} this repository holds. Unavailable this run: \texttt{T3\_indicsuper}.}
\label{tab:arms}
\end{table*}

\section{R\'enyi efficiency}
\label{sec:renyi}

R\'enyi efficiency \citep{zouhar-etal-2023-tokenization} scores a tokenizer by how evenly
its token-unigram distribution over a text uses the support that text actually touches, at
$\alpha > 1$ so the head of the distribution weighs more heavily than Shannon entropy
would. We normalise by the observed type count, following the authors' own reference
implementation; the variant normalised by nominal vocabulary size is stored alongside in
the results file. We report it as a diagnostic and nothing more.
\citet{cognetta-etal-2024-counterexamples} construct tokenizers whose R\'enyi efficiency
rises arbitrarily while the tokenization of the text is unchanged or worse, so on its own
the measure supports no claim here.

Table \ref{tab:renyi} reads as follows, at $\alpha = \numRenyiAlpha{}$ on the primary
prose corpus. The deployed arms cluster in a narrow band on Sanskrit,
\numRenyiDeployedLo--\numRenyiDeployedHi, with GPT-2 far below it at \numRenyiGptTwo. The
trained BPE arms sit above that band (\numRenyiBpeThirtyTwo{} at \numVocabSmall{} pieces
and \numRenyiBpeSixtyFour{} at \numVocabLarge) and the trained Unigram arms below it
(\numRenyiUnigramLo--\numRenyiUnigramHi). That comparison crosses columns: the trained
arms are scored in SLP1, which is their only variant, whereas the band just quoted is the
original-script column. The ordering also holds against the deployed arms' own SLP1
column, which runs \numRenyiDeployedSlpOneLo--\numRenyiDeployedSlpOneHi. The English side is lower than the Sanskrit side
for both deployed pivots scored on both halves of the same pairs, and across the pivots
and the matched controls it occupies a band of
\numRenyiEnglishLo--\numRenyiEnglishHi; the two Unigram Sanskrit arms sit at or just
below the foot of that band rather than above it. The reason this section carries no claim is visible in the same
table: R\'enyi ranks the two trained BPE arms in the opposite order from the controlled
tokens per proposition, putting the \numVocabSmall-piece arm above the
\numVocabLarge-piece one (\numRenyiBpeThirtyTwo{} against \numRenyiBpeSixtyFour) where the
controlled ratio prefers \numVocabLarge{} pieces (\numTppControlledBest{} against
\numTppControlledBpeThirtyTwo).

A further caution applies to the Sanskrit-against-English reading in particular.
Normalising by the support the text actually touches makes the measure sensitive to how
many tokens the text contains, and the two sides of the same pairs carry systematically
different token counts, which is this paper's whole finding. The Sanskrit-against-English
comparison in the paragraph above is therefore confounded by token count, and we state it
as an observation about the table rather than as a comparison between the languages.

For the deployed arms the original-script and SLP1 columns are the same text under the
same vocabulary, so the difference between them is what romanisation does to the token
distribution and nothing else. It lowers the measure for every deployed arm but GPT-2, to
\numRenyiDeployedSlpOneLo--\numRenyiDeployedSlpOneHi, and raises GPT-2's from
\numRenyiGptTwo{} to \numRenyiGptTwoSlpOne. We report the direction and do not interpret
it further.

\begin{table*}[t]
\centering
\footnotesize
\begin{tabular}{lrrrr}
\toprule
Arm & orig.\ $\alpha=2.5$ & orig.\ $\alpha=3.0$ & SLP1 $\alpha=2.5$ & SLP1 $\alpha=3.0$ \\
\midrule
\texttt{T0\_o200k} & 0.581 & 0.556 & 0.565 & 0.531 \\
\texttt{T0\_llama4} & 0.575 & 0.549 & 0.560 & 0.527 \\
\texttt{T0\_gemma3} & 0.568 & 0.536 & 0.563 & 0.529 \\
\texttt{T0\_gpt2} & 0.270 & 0.252 & 0.578 & 0.552 \\
\texttt{T3\_sarvam} & 0.570 & 0.541 & 0.556 & 0.530 \\
\texttt{T3\_sutra} & 0.580 & 0.549 & 0.525 & 0.490 \\
\texttt{T3\_brahmic131k} & 0.580 & 0.555 & 0.566 & 0.533 \\
\texttt{T1\_bpe\_raw\_32k}$^{*}$ & n/a & n/a & 0.615 & 0.564 \\
\texttt{T1\_bpe\_raw\_64k}$^{*}$ & n/a & n/a & 0.589 & 0.539 \\
\texttt{T2\_unigram\_raw\_32k}$^{*}$ & n/a & n/a & 0.493 & 0.456 \\
\texttt{T2\_unigram\_raw\_64k}$^{*}$ & n/a & n/a & 0.485 & 0.449 \\
\texttt{T0\_o200k} (English side) & n/a & n/a & 0.490 & 0.461 \\
\texttt{T0\_llama4} (English side) & n/a & n/a & 0.492 & 0.463 \\
\texttt{E1\_bpe\_32k} (English side) & n/a & n/a & 0.507 & 0.468 \\
\texttt{E1\_bpe\_64k} (English side) & n/a & n/a & 0.492 & 0.455 \\
\texttt{E1\_unigram\_32k} (English side) & n/a & n/a & 0.508 & 0.475 \\
\texttt{E1\_unigram\_64k} (English side) & n/a & n/a & 0.500 & 0.466 \\
\bottomrule
\end{tabular}
\caption{R\'enyi efficiency on S\=amayik test (prose), Sanskrit side in both script variants. The English-side rows carry the English text under the named pivot and have no script variant, so their values are placed in the right-hand pair of columns. Reported as a diagnostic only: the measure can be gamed, so it supports no claim in this paper on its own, and it is tabulated for the size-matched 32,000- and 64,000-piece arms and their pair-matched controls rather than extended over the byte-matched and 128,000-piece arms of Section~\ref{sec:rq2-controlled}.}
\label{tab:renyi}
\end{table*}

\section{The pre-registration}
\label{sec:preregistration}

\begin{table*}[t]
\centering
\small
\begin{tabular}{p{0.26\textwidth}p{0.42\textwidth}p{0.24\textwidth}}
\toprule
Pre-registered prediction & Measured & Outcome \\
\midrule
T0 fertility on Sanskrit 5--12 & 3.11--3.88 at $\geq$200k vocab; 12.49 for GPT-2 & Refuted, except GPT-2 on Devanagari \\
Hindi 2--4 under the same tokenizer & 1.79--2.34 at $\geq$200k vocab & Approximately held; lower bound crossed by 1 arm of 3 \\
Sa/Hi parity $>1.5$ & 1.060--1.353 & Refuted \\
T3 closes most of the gap vs Hindi & Sa/Hi 1.327--1.413 for T3 against 1.060--1.353 for T0 & Not observed; T3 arms sit within or above the T0 range \\
TPP crosses below 1.0 against \texttt{o200k}, for the proposed \texttt{T6} arm & \texttt{T6} is not built here. On the raw-subword baselines, the only Sanskrit-native arms this paper builds: 0.887 [0.875, 0.899] on prose against the deployed pivot; 1.035 [1.021, 1.049] under the matched control at 64,000 pieces, and 0.983 [0.971, 0.997] at 128,000 on in-domain prose against 1.025 [1.013, 1.037] out of domain & Observed against the pivot; under the matched control only at 128,000 in domain \\
\bottomrule
\end{tabular}
\caption{The predictions of our design document against what was measured. They were written before any arm was run and are dated only by our repository's history: this is a design document we did not edit, not a registration with an external timestamp, and it is reported for that reason and no stronger one. The first four are guesses at magnitudes rather than substantive hypotheses. The fifth was written for a sandhi-split, morpheme-constrained arm this paper does not build, so its Outcome is recorded against the raw-subword baselines this paper does build instead: the crossing is real against the deployed English pivot, and against the matched English control it appears only at the largest vocabulary trained here and only on in-domain prose.}
\label{tab:prereg}
\end{table*}

Table \ref{tab:prereg} carries the predictions of our design document, written before any
arm was run, against what was measured. They are reproduced unedited. The first four are
guesses at magnitudes rather than substantive hypotheses, and refuting one's own numeric
guesses is a statement about the guesses. The fifth was written for the sandhi-split,
morpheme-constrained arm this paper does not build, so its Measured column and its Outcome
carry the raw-subword baselines this paper does build instead: the crossing is real against
the deployed English pivot, and under the matched control it appears only at
\numVocabHuge{} pieces and only on in-domain prose. The document is dated only by our repository's history and carries
no external timestamp, so it is reported for what it is: a design we did not revise after
seeing the numbers.

\end{document}